%% file: Paper.tex
\documentclass[10pt,journal,compsoc]{IEEEtran}
\ifCLASSOPTIONcompsoc
\usepackage[nocompress]{cite}
\else
\usepackage{cite}
\fi
\ifCLASSINFOpdf
\else
\fi
\usepackage{enumerate}
\usepackage{color}
\usepackage[ruled,linesnumbered]{algorithm2e} 
\usepackage{amsmath}
\usepackage{multirow}
\usepackage{stfloats}
\usepackage{url}
\usepackage{graphicx}
\usepackage{xspace}
\usepackage{amssymb}
\usepackage{subfigure}

\newcommand*{\model}{{MVGCN}\@\xspace}

\gdef\eg{\textit{e.g.}}
\gdef\ie{\textit{i.e.}}
\gdef\etc{\textit{etc.}}
\gdef\G{\mathcal{G}}
\gdef\V{\mathcal{V}}
\gdef\E{\mathcal{E}}
\gdef\I{\mathbf{I}}
\gdef\O{\mathbf{O}}
\gdef\X{\mathbf{X}}
\gdef\W{\mathbf{W}}
\gdef\R{\mathbb{R}}
\newtheorem{problem}{Problem}
\newtheorem{definition}{Definition}
\graphicspath{{../../3-Figure/}{../../3-Figure/old/}}
\usepackage{amsmath}
\DeclareMathOperator*{\sigmoid}{sigmoid}
\DeclareMathOperator*{\concat}{concat}
\DeclareMathOperator*{\argmin}{arg\,min}

\makeatletter
\def\hlinew#1{%
	\noalign{\ifnum0=`}\fi\hrule \@height #1 \futurelet
	\reserved@a\@xhline}

\begin{document}
%
\title{Predicting Citywide Crowd Flows in Irregular Regions
	 Using Multi-View Graph Convolutional Networks}
%
%
%

\author{Junkai~Sun,
		Junbo~Zhang,~\IEEEmembership{Member,~IEEE,}
        Qiaofei~Li,
        Xiuwen~Yi,
        Yuxuan~Liang,
        Yu~Zheng,~\IEEEmembership{Senior Member,~IEEE}
\IEEEcompsocitemizethanks{
	\IEEEcompsocthanksitem J.K.~Sun, J.B.~Zhang, Y.~Zheng and X.W.~Yi are with JD Intelligent Cities Research, China and JD Intelligent Cities Business Unit, JD Digits, Beijing China. J.B.~Zhang and Y.~Zheng are also affiliated with Institute of Artificial Intelligence, Southwest Jiaotong University, Chengdu 611756, China. J.K.~Sun and Y.~Zheng are also affiliated with Xidian University, Xi'an 710071, China. E-mail: \{junkaisun, msjunbozhang, msyuzheng\}@outlook.com, xiuwenyi@foxmail.com.
	\IEEEcompsocthanksitem Q.F.~Li is with the School of Computer Science and Technology, Xidian University, Xi'an, 710071, China. E-mail: qjliqiaofei@gmail.com. 
	\IEEEcompsocthanksitem Y.X.~Liang is with the School of Computing, National University of Singapore. E-mail: yuxliang@outlook.com.
}
\thanks{Junbo Zhang and Junkai Sun contributed equally to this work.}
\thanks{Junbo Zhang and Yu Zheng are the corresponding authors.}

}

%
%

\markboth{Journal of \LaTeX\ Class Files,~Vol.~14, No.~8, August~2015}%
{Shell \MakeLowercase{\textit{et al.}}: Bare Demo of IEEEtran.cls for Computer Society Journals}
%



\IEEEtitleabstractindextext{%
	\begin{abstract}
		
		Being able to predict the crowd flows in each and every part of a city, especially in \textit{irregular regions}, is strategically important for traffic control, risk assessment, and public safety. However, it is very challenging because of interactions and spatial correlations between different regions. In addition, it is affected by many factors: i) multiple \textit{temporal correlations} among different time intervals: closeness, period, trend; ii) complex \textit{external} influential factors: weather, events; iii) \textit{meta} features: time of the day, day of the week, and so on. 
		
		In this paper, we formulate crowd flow forecasting in irregular regions as a \textit{spatio-temporal graph} (STG) prediction problem in which each node represents a region with time-varying flows. By extending \textit{graph convolution} to handle the spatial information, we propose using \textit{spatial graph convolution} to build a \textit{multi-view graph convolutional network} (MVGCN) for the crowd flow forecasting problem, where different views can capture different factors as mentioned above. We evaluate MVGCN using four real-world datasets (taxicabs and bikes) and extensive experimental results show that our approach outperforms the adaptations of state-of-the-art methods. And we have developed a crowd flow forecasting system for irregular regions that can now be used internally. 
	\end{abstract}
	
}

\maketitle
\input{introduction}
\input{preliminaries}

\input{method}

\input{experiments}

\input{relatedwork}

%
\IEEEpeerreviewmaketitle

\section*{Acknowledgments}


This work was supported by the National Key R\&D Program of China (2019YFB2101805), Beijing Academy of Artificial Intelligence (BAAI), and the National Natural Science Foundation of China (Grant No. 61672399). 

\ifCLASSOPTIONcaptionsoff
  \newpage
\fi



%
%
%

\bibliographystyle{IEEEtranS}
\bibliography{ref.bib}

%

\begin{IEEEbiography}[{\includegraphics[width=1in,height=1.25in,clip,keepaspectratio]{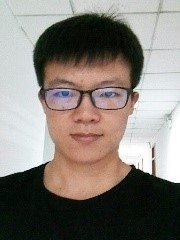}}]{Junkai Sun}
	is a Master's student in Xidian University, majoring in computer science and technology. His research interests mainly include spatio-temporal data mining with deep learning, and urban computing. He is also an intern at JD Intelligent Cities Business Unit, JD Digits. 
\end{IEEEbiography}

\begin{IEEEbiography}[{\includegraphics[width=1in,height=1.25in,clip,keepaspectratio]{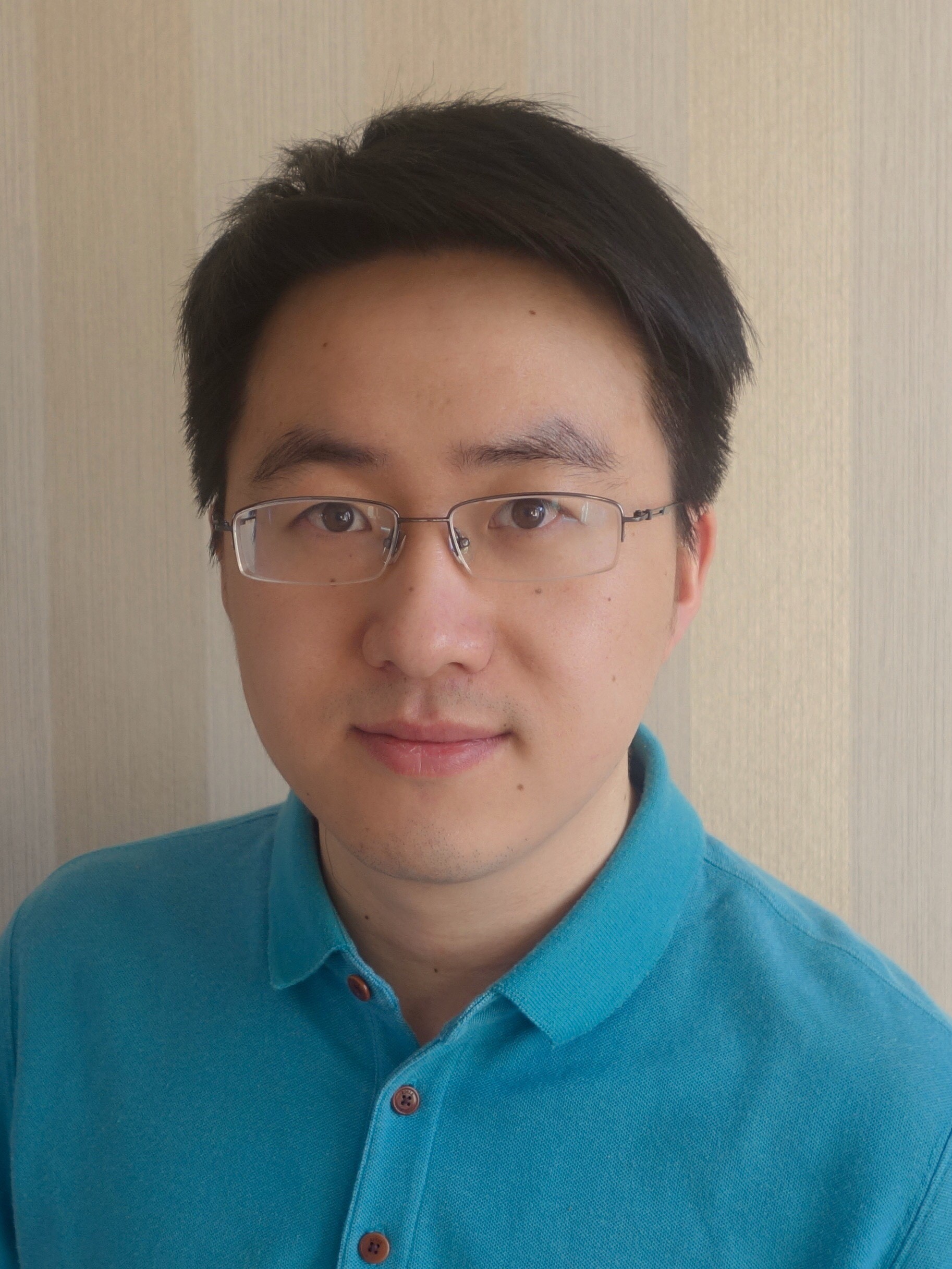}}]{Dr.~Junbo Zhang}
	is a Senior Researcher of JD Intelligent Cities Research and the Head of AI Department, JD Intelligent Cities Business Unit, JD Digits. Prior to that, he was a researcher at MSRA from 2015 - 2018. His research interests include urban computing, machine learning, data mining, and big data analytics. He currently serves as Associate Editor of ACM Transactions on Intelligent Systems and Technology. He has published over 30 research papers (e.g. AI Journal, IEEE TKDE, KDD, AAAI, IJCAI) in refereed journals and conferences, among which one paper was selected as the ESI Hot Paper, three as the ESI Highly Cited Paper. Dr. Zhang received the ACM Chengdu Doctoral Dissertation Award in 2016, the Chinese Association for Artificial Intelligence (CAAI) Excellent Doctoral Dissertation Nomination Award in 2016, the Si Shi Yang Hua Medal (Top $1/1000$) of SWJTU in 2012, and the Outstanding Ph.D. Graduate of Sichuan Province in 2013. He is a member of IEEE, ACM, CAAI and China Computer Federation.
\end{IEEEbiography}

\begin{IEEEbiography}[{\includegraphics[width=1in,height=1.25in,clip,keepaspectratio]{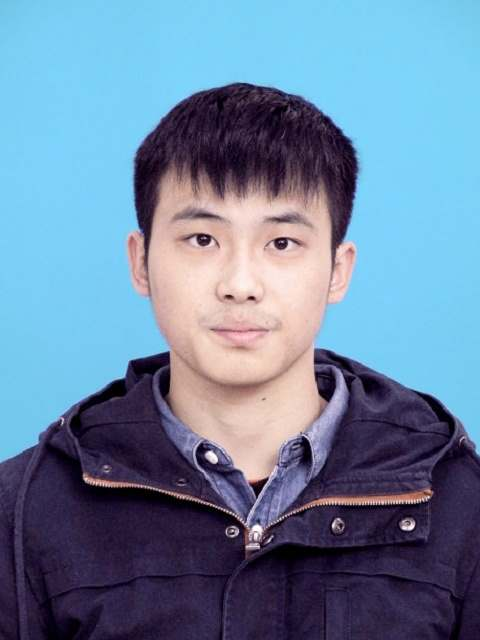}}]{Qiaofei Li}
	is a Master's student in Xidian University, majoring in computer science and technology. His research interests focus on deep learning and spatio-temporal data mining.
\end{IEEEbiography}

\begin{IEEEbiography}[{\includegraphics[width=1in,height=1.25in,clip,keepaspectratio]{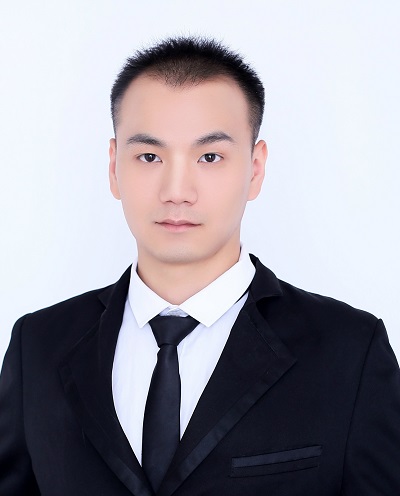}}]{Dr.~Xiuwen Yi}
	is currently a Data Scientist of JD Intelligent Cities Business Unit, JD Digits, focuses on using big data and AI technology to build real-world applications for tackling urban challenges. He got his Ph.D. degree in Computer Science and Technology from Southwest Jiaotong University in 2018. He was an intern in Urban Computing Group at MSR Asia from 2014 to 2017. His research interests include: Spatiotemporal Data Mining, Deep Learning, and Urban Computing. He has published over 15 research papers in refereed conferences (e.g., KDD, IJCAI) and journals (e.g., EEE TKDE, AI). 
\end{IEEEbiography}

\begin{IEEEbiography}[{\includegraphics[width=1in,height=1.25in,clip,keepaspectratio]{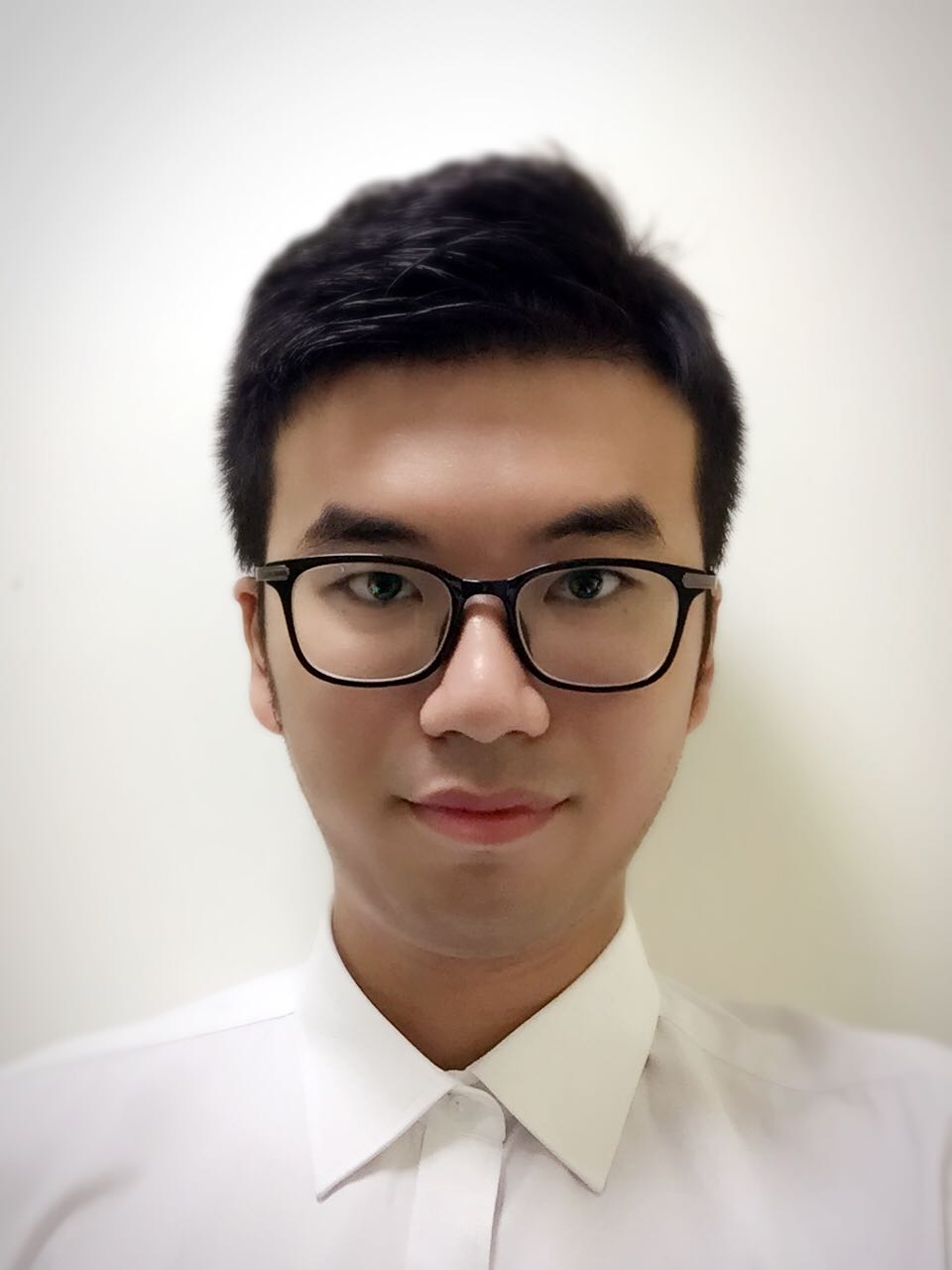}}]{Liang}
is currently pursuing his Ph. D. degree at School of Computing, National University of Singapore. He has published several papers in refereed conferences, such as KDD, IJCAI and AAAI. His research interests mainly lie in machine learning, deep learning and their applications in urban areas.
\end{IEEEbiography}

\begin{IEEEbiography}[{\includegraphics[width=1in,height=1.25in,clip,keepaspectratio]{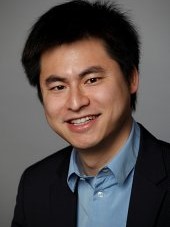}}]{Dr.~Yu Zheng}
	is a Vice President of JD.COM and the Chief Data Scientist of JD Digits. He also leads the JD Intelligent Cities Business Unit as the president and serves as the managing director of JD Intelligent Cities Research. He is also a Chair Professor at Shanghai Jiao Tong University and an Adjunct Professor at Hong Kong University of Science and Technology and Nanjing University. Before joining JD Digits, he was a senior research manager at Microsoft Research. Zheng currently serves as the Editor-in-Chief of ACM Transactions on Intelligent Systems and Technology. He has served as chair on over 10 prestigious international conferences, e.g. as the program co-chair of ICDE 2014 (Industrial Track) and CIKM 2017 (Industrial Track). In 2013, he was named one of the Top Innovators under 35 by MIT Technology Review (TR35) and featured by Time Magazine for his research on urban computing. In 2014, he was named one of the Top 40 Business Elites under 40 in China by Fortune Magazine. In 2017, Zheng was named an ACM Distinguished Scientist. 
\end{IEEEbiography}



\end{document}

%% file: introduction.tex
\section{Introduction}
%
%
%
%
\IEEEPARstart{F}{orecasting} crowd flows in each and every part of a city, especially in \textit{irregular regions}, plays an important role in traffic control, risk assessment, and public safety. For example, when vast amounts of people streamed into a strip region at the 2015 New Year's Eve celebrations in Shanghai, this resulted in a catastrophic stampede that killed 36 people. Such tragedies can be mitigated or prevented by utilizing emergency mechanisms, like sending out warnings or evacuating people in advance, if we can accurately forecast the crowd flow in a region ahead of time. 

\begin{figure}[!htbp]
	\centering
	\includegraphics[width=.8\linewidth]{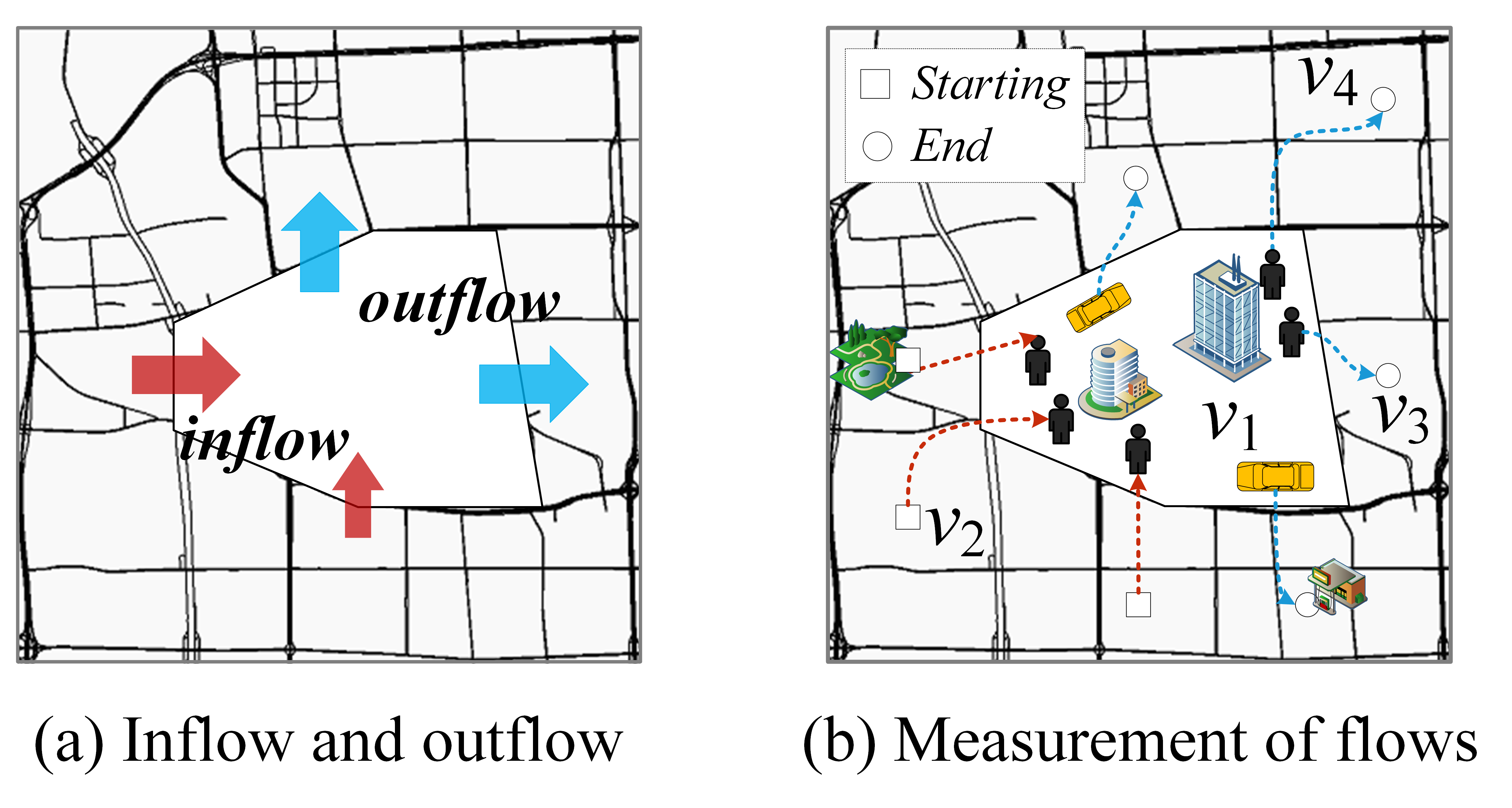}
	\vspace*{-12pt}
	\caption{Crowd flows in an irregular region}
	\label{fig:crowd_flows}
\end{figure}

Prior works mainly focused on predicting the crowd flows in \textit{regular gridded regions} \cite{zhang2017aaai,zhang2019multitask,yao2019revisiting}.
Although partitioning a city into grids is more easily and effectively handled by the subsequent  data mining \cite{Zheng2014AToISaTT} and machine learning approaches \cite{zhang2017aaai}, the regions in a city are actually separated by road networks and therefore extremely \textit{irregular}. There are also some existing literature that models the non-Euclidean correlation using graph techniques to in the forecasting problems \cite{geng2019multigcn, lin2018air,zhang2018gan}. Different from these previous attempts, our work consists of three tasks: data preprocessing, map segmentation and traffic forecasting. It first takes the raw trajectories and road networks as inputs ans then simultaneously considers multi-view temporal features as well as external views.
In this study, our goal is to collectively predict inflow and outflow of crowds in each and every \textit{irregular} region of a city. 
Fig~\ref{fig:crowd_flows} shows an illustration. \textit{Inflow} is the total flow of crowds entering a region from other regions during a given time interval and \textit{outflow} denotes the total flow of crowds leaving a region for other regions during a given time interval, both of which track the transition of crowds between regions. Knowing them is very beneficial for traffic control. 

We can measure crowd flows by the number of cars/bikes running on the roads, the number of pedestrians, the number of people traveling on public transportation systems (\eg{} metro, bus), or all of them together if the data is available. 
We can use the GPS trajectories of vehicles to measure the traffic flow, showing that the inflow and outflow of $v_1$ are (0, 2) respectively. 
Similarly, using mobile phone signals of pedestrians, the two types of flows are (3, 2) respectively.

\begin{figure*}[b]
	\centering
	\includegraphics[width=.99\linewidth]{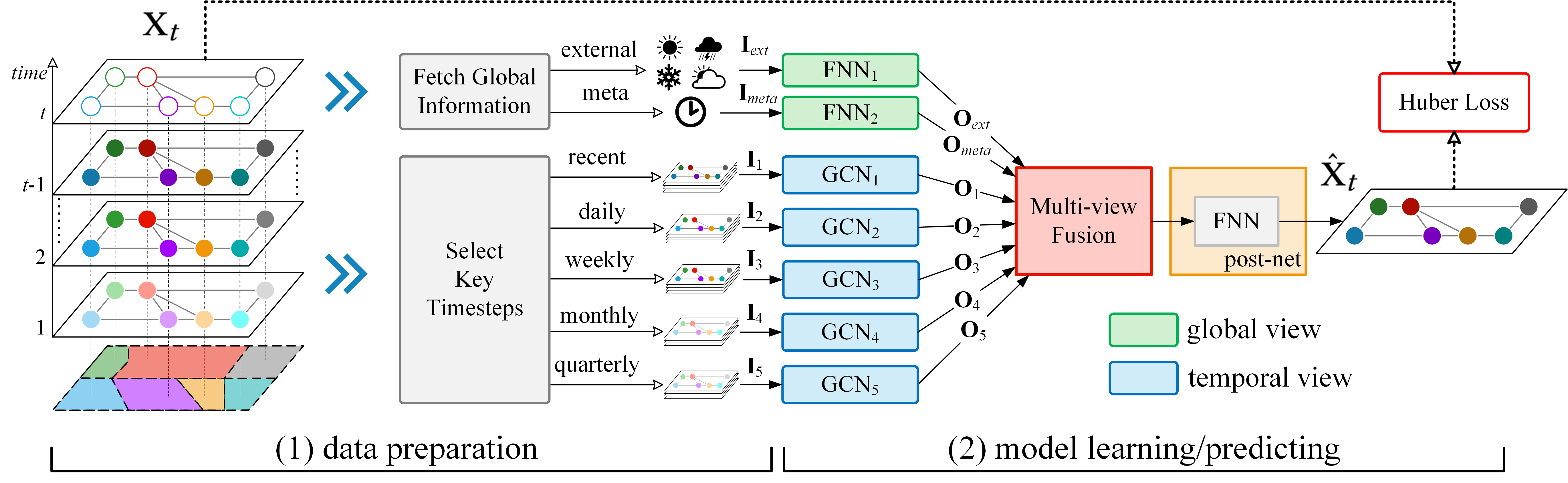}
	\vspace*{-6pt}
	\caption{Multi-view deep learning framework. (1) Data preparation stage: fetching global information based on the predicted target time and selecting key timesteps based on temporal dependencies. (2) Model learning stage: a) Graph convolutional net (GCN) is used to learn the spatial correlations and interactions using the structural information of the STG; b) Fully-connected nueral net (FNN) is employed to capture global information, like external factors and meta features (time of the day). c) Multi-view fusion can effectively integrate the outputs of GCNs and FNNs. d) Post-net, namely a FNN here, is used to project the latent representation to the output using an activation function (\eg{} $\tanh$). }
	\label{fig:arch}
\end{figure*}

We formulate the crowd flow forecasting problem as a spatio-temporal graph (STG) prediction problem in which an \textit{irregular region} can be viewed as a \textit{node} that is associated with time-varying inflow and outflow, and transition flow between regions can be used to construct the edges. 
However, forecasting these two kinds of flows in each and every node of an STG is very challenging because of the following three complex factors: 

\noindent1)~\textbf{Interactions and spatial correlations between different vertices of an STG}. The inflow of the node $v_1$ (Fig~\ref{fig:crowd_flows}(b)) is affected by outflows of \textit{adjacent} (1-hop) neighbors ($v_2$ and $v_3$) as well as \textit{multi-hop} neighbors (other nodes, like $v_4$). Likewise, a node's outflow would affect its neighbors' inflows. Moreover, a node's inflow and outflow interact with each other.

\noindent2)~\textbf{Multiple types of temporal correlations among different time intervals}: closeness, period, and trend. i) \textit{Closeness}: the flows of a node are affected by recent time intervals. Taking traffic flow as an example, a congestion occurring at 5pm will affect traffic flow at 6pm. ii) Many types of \textit{Periods}: daily, weekly, \textit{etc}. Traffic conditions during rush hours may be similar on consecutive workdays (daily period) and consecutive weekends (weekly period). iii) Many types of \textit{Trends}: monthly, quarterly, \textit{etc}. Morning peak hours may gradually happen earlier as summer comes, with people getting up earlier as the temperature gradually increases and the sun rises earlier. 

\noindent3)~\textbf{Complex external factors and meta features}. Holidays can influence the flow of crowds for consecutive days, and extreme weather always changes the crowd flows tremendously in different regions of a city. Besides, crowd flows are also affected by meta data, like time of day, weekend/weekday. For example, the flow patterns on rush hours may differ from non-rush hours. 

To tackle all aforementioned challenges, we propose a general multi-view learning framework for  crowd flow prediction in all the irregular regions of a city, as shown in Fig~\ref{fig:arch}.
The framework is composed of two stages: data preparation and model learning. The data preparation stage involves fetching global information based on the target time and selecting the dependent crowd flow matrices from key timesteps according to different temporal properties. 
Based on the collected multiple view data, we present a new model for learning, which we refer to as a \textit{multi-view graph convolutional network} (MVGCN), consisting of several GCNs and fully-connected neural networks (FNNs). 
The contributions of this research lie in the following four aspects: 

\begin{itemize}
 \item We propose a variant of GCN, which can capture spatial correlations between different nodes. We design a multi-view fusion module, to fuse multiple latent representations from different views.The module is designed based on two fusion methods: gating and sum fusion, which are used to capture sudden and slight changes, respectively. 
	
\item We propose a comprehensive framework that consists of data preprocessing, map segmentation and map clustering by road networks, graph construction via transition flows, crowd flow prediction using graph convolutional networks. Besides, we also design a demo system to visualize the crowd flow forecasting results in citywide irregular regions.
	
\item We evaluate our \model{} using four real-world mobility datasets, including taxicab data in Beijing and New York City (NYC), and bike data in NYC and Washington D.C. The extensive results demonstrate advantages of our \model{} beyond the adaptations of several state-of-the-art approaches, like diffusion convolutional recurrent neural networks \cite{lidiffusion} and Gaussian Markov random field based model \cite{hoang2016fccf}. 
\end{itemize}

%% file: preliminaries.tex
\section{Problem Definition}

\subsection{Irregular Regions}\label{sec:region}
Urban areas are naturally divided into different irregular regions by road network. These regions may have different functions, such as education and business function \cite{yuan2012}. Different functional areas usually have different traffic flow patterns. For example, most people usually commute from residential areas to work places in the morning and return home after work. So it is actually more rational and insightful to perform the task of traffic flow prediction on these irregular regions. 

\begin{figure}[!htbp]
	\centering
	{\includegraphics[width=0.99\linewidth]{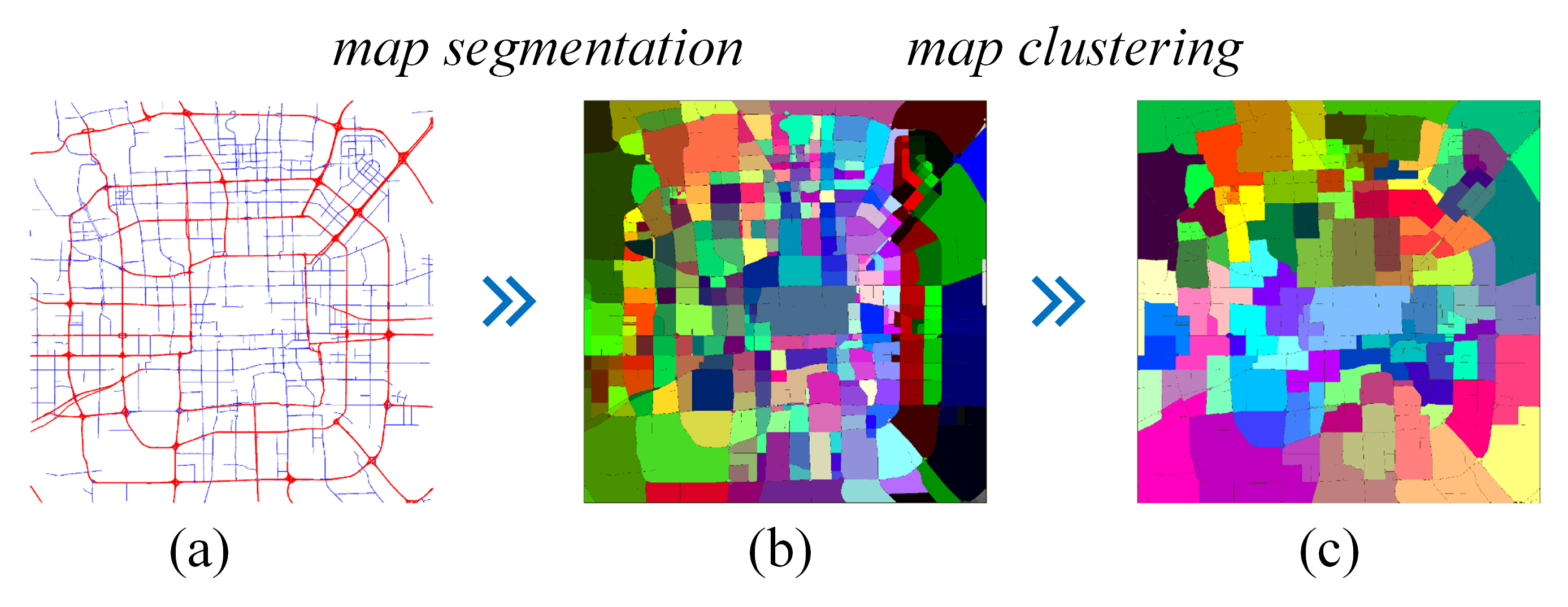}}
	\vspace*{-12pt}
	\caption{(a) Road network in Beijing; (b) Regions after map segmentation; (c) Regions after map clustering. }
	\label{road_network}
\end{figure}

\noindent\textbf{Region Partition}. 
The task of \textit{region partition} consists of two main operations: map segmentation and map clustering. For example, the road network in Beijing is composed of multi-level roads, such as level 0, 1, 2, etc., which represent different functional road categories. As shown in Figure \ref{road_network} (a), the red segments denote highways and city express ways, and the blue segments represent urban arterial roads in Beijing.

\begin{figure}[!htbp]
	\centering
	{\includegraphics[width=0.99\linewidth]{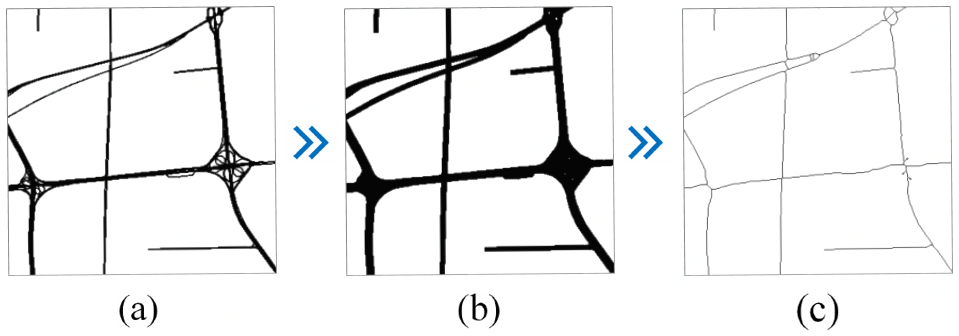}}
	\vspace*{-12pt}
	\caption{(a) Before Dilation; (b) After Dilation; (c) After Thinning. }
	\label{image_process}
\end{figure}

Referring to \cite{yuan2012segmentation}, we utilize morphological image processing techniques to tackle the region partition task. Specifically, we partition the map into $2400\times2400$ small grid-cells, and map each road point to its corresponding grid-cells, thereby obtaining a binary image, in which 1 and 0 stand for road segments and blank areas respectively. Then we apply the dilation operation and thinning operation to get the skeleton of the road network. The dilation operation can help thicken the roads, fill the small holes and smooth out unnecessary details. Then the thinning operation is used to recover the size of a region while keeping the connectivity between region, as shown in Figure \ref{image_process}. Finally, we can obtain all labeled irregular regions' locations using the connected component labeling algorithm (CCL) that finds individual regions by clustering "1"-labeled grids.

After the map segmentation, we obtain a large number of low-level irregular regions, and many of them are too small to collect or predict traffic flows at the city scale. Therefore, we apply a clustering operation \cite{KarypisK99} on these regions. Specifically, we define the edge weight between two low-level regions as the Spearman's rank correlation coefficient between the average crowd flows within a time period (\eg{} one day). After this operation, the small intractable regions are clustered into some high-level regions, as shown in Figure \ref{road_network} (b) and (c). 


\noindent\textbf{Graph Construction}. 
To capture the spatial dependency of traffic flow between different irregular regions, we construct a topological graph using historical region-wise transition flow. The intuition is that adjacent regions in geo-space are usually closely correlated, besides that, regions that are distant can also influence each other due to the convenient transportation such as subway, taxi and so on. Transition flow can reflect the traffic interaction between close or distant regions. 
Specifically, we select a period of time from the traffic data, such as one or two months. Then we can statistic the valid time slices between pair-wise regions. Valid time slice means when the region-wise transition flow is greater than a threshold $\alpha$ considering the noise of trajectories data. When the ratio of valid time slices for region-wise transition is greater than a threshold $\beta$, we place an binary value undirected edge to connect them. In our paper, the thresholds $\alpha$ is set as 3, $\beta$ is set as 0.1.

\begin{table}[b]\fontsize{9}{12}\selectfont
	\centering
	\caption{Notation}
	\vspace*{-5pt}
	\label{tab:notation}
	\begin{tabular}{l|l}
		\hline
		Symbol                                                    & Description                                                 \\ \hline
		$\mathcal G=(\V,\E)$ & spatio-temporal graph \\
		
		$\V=\{v_i\}$ & a set of $N$ nodes, $i=1,\cdots, N$\\
		$\mathbf A \in \mathbb R^{N\times N}$ & an adjacency matrix \\
		$\mathbf S \in \mathbb R^{N\times N}$ & a modified adjacency matrix \\
		$P=\{p_i\}$ & geospatial position of node $v_i$ \\
		
		$\mathcal T$ & available time interval set \\
		$\mathbf X_t \in \mathbb R^{N\times C} $ & a matrix of node feature vectors at $t\in \mathcal T$                                     \\
		$\mathbf X_t[i,:]$                   & vector of node $n_{i}$                                     \\
		$\mathbf X_t[:,c]$                   & vector of $c$-th channel in all nodes                                   \\
		\hline
	\end{tabular}
\end{table}

\subsection{Prediction Problem on Spatio-Temporal Graphs} 
The goal in this research is to collectively predict the future inflows/outflows in each and every node of an STG based on historical observations. 
Table~\ref{tab:notation} lists the mathematical notation used in the paper. 

\begin{definition}[STG]
	A spatio-temporal graph (STG), denoted as $\mathcal G=(\mathcal V, \mathcal E, \mathbf A)$, where $\mathcal V$ and $\mathcal E$ respectively denote the set of $|\V|=N$ vertices and edges, $\mathbf A \in \mathbb R^{N\times N}$ is a binary unweighted adjacency matrix. Specifically, each vertex $v_i \in \V$ has a geospatial position $p_i$ and time-varying attributes. These attributes over an STG at time $t$ can be viewed as a graph signal $\X_t \in \mathbb R^{N \times C}$, where $\X_t[i,:]\in \R^C$ represents $C$ attributes in the node $v_i$, \eg, the inflow and outflow \cite{zhang2017aaai} ($C=2$). The edges between two regions are constructed from region-wise transition flows and the binary entry value in $\mathbf A$ indicates whether two regions are correlated in traffic flow.
\end{definition}
\begin{problem}
	Given a graph $\mathcal G=(\mathcal V, \mathcal E, \mathbf A)$ and observed attributes of nodes $\{\mathbf X_{t}|t=1,2,\cdots,T\}$, predict the attributes at the next time step, \ie, $\mathbf X_{T+1}$. 
\end{problem}

%% file: method.tex
\section{Methodology}
In this section, we present our new model for crowd flow forecasting. 
We first present a multi-view deep learning framework \cite{xu2013survey, zhao2017survey}, then we review the graph convolutional network and present our new spatial graph convolutional network. Finally, we present the multi-view fusion method and loss function used in our model. 
\subsection{Multi-view Deep Learning Framework}\label{sec:framework}
Fig~\ref{fig:arch} provides an overview of our proposed deep learning framework to predict the crowd flows in an STG. 
We adopt the multi-view framework that is an effective mechanism to learn latent representations from cross domain data \cite{wang2015deep}. 
The framework proposed is composed of two stages: data preparation and model learning/predicting. The first stage is used to fetch global information and select the key timesteps, then we feed all of them to the second stage to perform model training. We provide concrete details in the following sections.

\vspace*{4pt}
\noindent\textbf{Data preparation stage}. ``What factors should be considered when forecasting the crowd flow in a region?'' a) weather, b) time of the day, c) period, \etc{} Different people may have different answers that highlights different views on this problem. We summarize these views into two categories: \textit{global view} and \textit{temporal view}. 
(1) the global view is composed of external and meta views. 
According to the time of the predicted target, we fetch different external data, like meteorological data in previous timesteps and weather forecasting. We can also construct the meta features: time of the day, day of the week, and so on. The external and meta features are represented as $\I _{ext}$ and $\I _{meta}$, respectively. 
(2) the temporal view contains multiple views according to the temporal closeness, period, trend. Considering two types of periods (daily and weekly), and two types of trends (monthly and quarterly)\footnote{One can set different periods and trends in practice, like yearly period, based on the characteristics of the data.}, we select the corresponding recent, daily, weekly, monthly, and quarterly timesteps as the key timesteps, to construct five views. 
For each of the different temporal views, we fetch a list of key timesteps' flow matrices and concatenated them, to construct five inputs as follows, 
\begin{align}
&\mathbf I_1 = \concat[\mathbf X_{t-1}, \mathbf X_{t-2}, \cdots, \mathbf X_{t-l_r}] \in \mathbb R^{N\times C\times l_r} \nonumber \\
&\mathbf I_2 = \concat[\mathbf X_{t-p_d}, \mathbf X_{t-2p_d}, \cdots, \mathbf X_{t-l_d*p_d}] \in \mathbb R^{N\times C\times l_d} \nonumber \\
&\mathbf I_3 = \concat[\mathbf X_{t-p_w}, \mathbf X_{t-2p_w}, \cdots, \mathbf X_{t-l_w*p_w}] \in \mathbb R^{N\times C\times l_w} \nonumber\\
&\mathbf I_4 = \concat[\mathbf X_{t-p_m}, \mathbf X_{t-2p_m}, \cdots, \mathbf X_{t-l_m * p_m}] \in \mathbb R^{N\times C\times l_m} \nonumber\\
&\mathbf I_5 = \concat[\mathbf X_{t-p_q}, \mathbf X_{t-2p_q}, \cdots, \mathbf X_{t-l_q * p_q}] \in \mathbb R^{N\times C\times l_q} \nonumber
\end{align}
where $l_r$, $l_d$, $l_w$, $l_m$, and $l_q$ are input lengths of recent, daily, weekly, monthly, and quarterly lists, respectively. $p_d$ and $p_w$ are daily and weekly periods; $p_m$ and $p_q$ are monthly and quarterly trend spans. 

By selecting these key timesteps, our approach can capture multiple types of temporal properties. The complexity of the input data of our approach is $l_r + l_d + l_w + l_m + l_q$, and these views can be modeled in parallel. 
If one uses a sequence neural network model (like recurrent neural networks, RNNs) to capture all these temporal dependencies automatically, the complexity would be $O(\max(l_r, l_d*p_d, l_w*p_w, l_m * p_m, l_q * p_q)) = O(l_q * p_q)$, while RNNs maintain a hidden state of the entire past that prevents parallel computation within a sequence. Assuming lengths of recent, daily, weekly, monthly, and quarterly lists are all equal 3, our architecture only needs $3 \times 5 = 15$ key frames. In contrast, RNNs needs 3 quarters of data, approximately $24\ \rm frames/day\ \times 30\ days/month\ \times 3\ months/quarter\ \times 3\ quarters = 6480$ frames. Such a long-range sequence tremendously raises the training complexity for RNNs, making them infeasible in real-world applications.  

\noindent\textbf{Model learning/predicting stage}. We employ graph convolutional networks (GCNs, see Section~\ref{sec:gcn}) and fully-connected neural networks (FNNs) to model the temporal and global views, respectively. For each temporal view, GCN is used to learn the time-varying spatial correlations and interactions using the structural information of the STG, and The corresponding outputs of five GCNs are denoted $\mathbf O_1,\cdots, \mathbf O_5 \in \mathbb R^{N\times C}$. Two FNNs are employed to capture the influences from external and meta data, respectively, and the outputs are $\O_{ext}$ and $\O_{meta}$. All these outputs are then fed into the \textit{multi-view fusion module} (see Section~\ref{sec:mvf}) followed by a post-net (\eg{} FNN), to obtain the final prediction $\hat\X_t$. The multi-view fusion can effectively employ the outputs of different views based on their characteristics. Finally, we apply the Huber loss\cite{huber1964robust} for robust regression.

\subsection{Graph Convolutional Network for STG}\label{sec:gcn}
\noindent\textbf{Convolutional networks over graphs}. 
Recently, generalizing convolutional networks to graphs have become an area of interest. 
In this paper, we mainly consider spectral convolutions \cite{bruna2014spectral,defferrard2016convolutional} on arbitrary graphs.
As it is difficult to express a meaningful translation operator in the node domain \cite{bruna2014spectral}, \cite{defferrard2016convolutional} presented a spectral formulation for the convolution operator on the graph, denoted as $*_\G$. By this definition, the graph signal $\X \in \mathbb R^{N\times C}$ 
with a filter $g_\textbf{w} = \rm{diag}(\textbf{w})$ parameterized by $\textbf{w} \in \R^{N}$ in the Fourier domain, 
\begin{eqnarray}
g_\textbf{w} *_\G \mathbf X = g_\textbf{w} (\mathbf L)\mathbf X = g_\textbf{w} (\mathbf U \Lambda \mathbf U ^\top) \mathbf X \label{eq:spectral}
\end{eqnarray}
where $\textbf{U} \in \mathbb R^{N\times N}$ is the matrix of eigenvectors, and $\Lambda \in \mathbb R^{N\times N}$ is the diagonal matrix of eigenvalues of the normalized graph Laplacian $\textbf{L}=\textbf{I}_N - {\textbf{D}}^{-\frac{1}{2}} {\textbf{A}} {\textbf{D}}^{-\frac{1}{2}} = \textbf{U} \Lambda \textbf{U}^\top \in \mathbb R^{N\times N}$, where $\textbf{I}_N$ is the identity matrix and $\textbf{D} \in \mathbb R^{N\times N} $ is the diagonal degree matrix with $\textbf{D}_{ii}=\sum_j{\textbf{A}}_{ij}$. We can understand $g_\textbf{w}$ as a function of the eigenvalues of $\mathbf L$. However, evaluating Eq.~\ref{eq:spectral} is computationally expensive, as the multiplication with $\textbf{U}$ is $\mathcal O(N^2)$. 
To circumvent this problem, the Chebyshev polynomial expansion (up to $K^{th}$ order) \cite{defferrard2016convolutional} was applied to obtain an efficient approximation, as
\begin{eqnarray}
g_\textbf{w} (\mathbf L)\mathbf X \approx \sum_{k=0}^{K-1}\textbf{w}_k (\mathbf L^k)\mathbf X = \sum_{k=0}^{K-1}\textbf{w}_k'T_k(\tilde{\textbf L})\mathbf X\label{eq:Cheb}
\end{eqnarray}
where $T_k(\tilde{\textbf L})$ is the Chebyshev polynomial of order $k$ evaluated at the scaled Laplacian $\tilde{\textbf L} = {\frac{2}{\lambda_{\max}}} \textbf{L} - \textbf{I}_N$, $\lambda_{\max}$ denotes the largest eigenvalue of $\textbf{L}$, $\textbf w' \in \mathbb R^K$ is now a vector of Chebyshev coefficients. The details of this approximation can be found in \cite{hammond2011wavelets,defferrard2016convolutional}. 

Furthermore, \cite{kipf2017semi} proposed a fast approximation of the spectral filter by setting $K=1$ and successfully used it for semi-supervised classification of nodes, as
\begin{eqnarray}
\textbf{Y} = \tilde{\textbf{D}}^{-\frac{1}{2}} \tilde{\textbf{A}} \tilde{\textbf{D}}^{-\frac{1}{2}} \X \W \label{eq:gconv}
\end{eqnarray}
where $\textbf{Y} \in \mathbb R^{N\times F}$ is the signal convolved matrix. $\tilde{\textbf{A}} = \textbf{A} + \textbf{I}_N$ is the adjacency matrix of $\G$ with added self-connections, $\tilde{\textbf{D}}_{ii}=\sum_j\tilde{\textbf{A}}_{ij}$ and $\W \in \mathbb R^{C\times F}$ is a trainable matrix of filter parameters in a graph convolutional layer. The filtering operation has complexity $\mathcal O(|\E|FC)$ as $\tilde{\textbf{A}}\X$ \cite{kipf2017semi} and can be efficiently implemented as the product of a sparse matrix with a dense matrix. 


\noindent\textbf{Spatial graph convolutional network}.
We present a variant of fast approximate graph convolution (Eq.~\ref{eq:gconv}) that also considers the geospatial positions of vectices in an STG. Here we explore an approach to integrate such geospatial positions based on the First Law of Geography \cite{tobler1970computer}, \ie{}, everything is related to everything else, but near things are more related than distant things. 

Given an adjacency matrix $\textbf{A}$, we assign spatial weights for existing edges based on the spatial distance, as
\begin{eqnarray}
\mathbf S = \mathbf A \odot \mathbf \omega
\end{eqnarray}
where $\textbf{S}\in \mathbb R^{N\times N}$ is the modified adjacency matrix, $\odot$ is the Hadamard product (\ie{} element-wise multiplication).
$\omega\in \mathbb R^{N\times N}$ is the spatial weighted adjacency matrix that is calculated via a thresholded Gaussian kernel weighting function \cite{shuman2013emerging}, as
\begin{eqnarray}
\omega_{ij} = \left\{
\begin{array}{lr}
\exp \left( -\frac{[dist(p_i,p_j)]^2}{2\theta ^2} \right) & {\rm if \,} {\rm dist}(p_i,p_j) \leq \kappa \\
0 & \rm{otherwise} \\
\end{array}
\right.
\end{eqnarray}
Here ${\rm dist}(p_i,p_j)$ means the geographical distance between nodes $v_i$ and $v_j$; $\theta$ and $\kappa$ are two parameters to control the scale and sparsity of the adjacency matrix. 

With the modified matrix $\mathbf S$, we consider multiple graph convolutional layers with the following layer-wise propagation rule:
\begin{eqnarray}
\textbf{H}^{(l+1)} = f\left( {\textbf{Q}}^{-\frac{1}{2}} \tilde{\textbf{S}} {\textbf{Q}}^{-\frac{1}{2}} \textbf{H}^{(l)} \W^{(l)}\right) \label{eq:sc}
\end{eqnarray}
where $\textbf{H}^{(l+1)} \in \mathbb R^{N\times F_{l+1}}$ and $\textbf{H}^{(l)} \in \mathbb R^{N\times F_{l}}$ are the output and input of the $l^{th}$ layer. $\tilde{\textbf{S}} = \textbf{S} + \textbf{I}_N$ is the modified adjacency matrix with added self-connections, $\textbf{Q}_{ii}=\sum_j\tilde{\textbf{S}}_{ij}$ and $\W^{(l)} \in \mathbb R^{F_{l}\times F_{l+1}}$ is a trainable matrix of filter parameters in a graph convolutional layer, $f$ denotes an activation function, \eg{} the rectifier $f(z) := \max(0, z)$ \cite{krizhevsky:2012imagenet}. The filtering operation has complexity $\mathcal O(|\E|F_lF_{l+1})$ as $\tilde{\textbf{S}}\mathbf H^{(l)}$ can be efficiently implemented as a product of a sparse matrix with a dense matrix. 

\begin{figure}[!bthp]
	\centering
	\includegraphics[width=.99\linewidth]{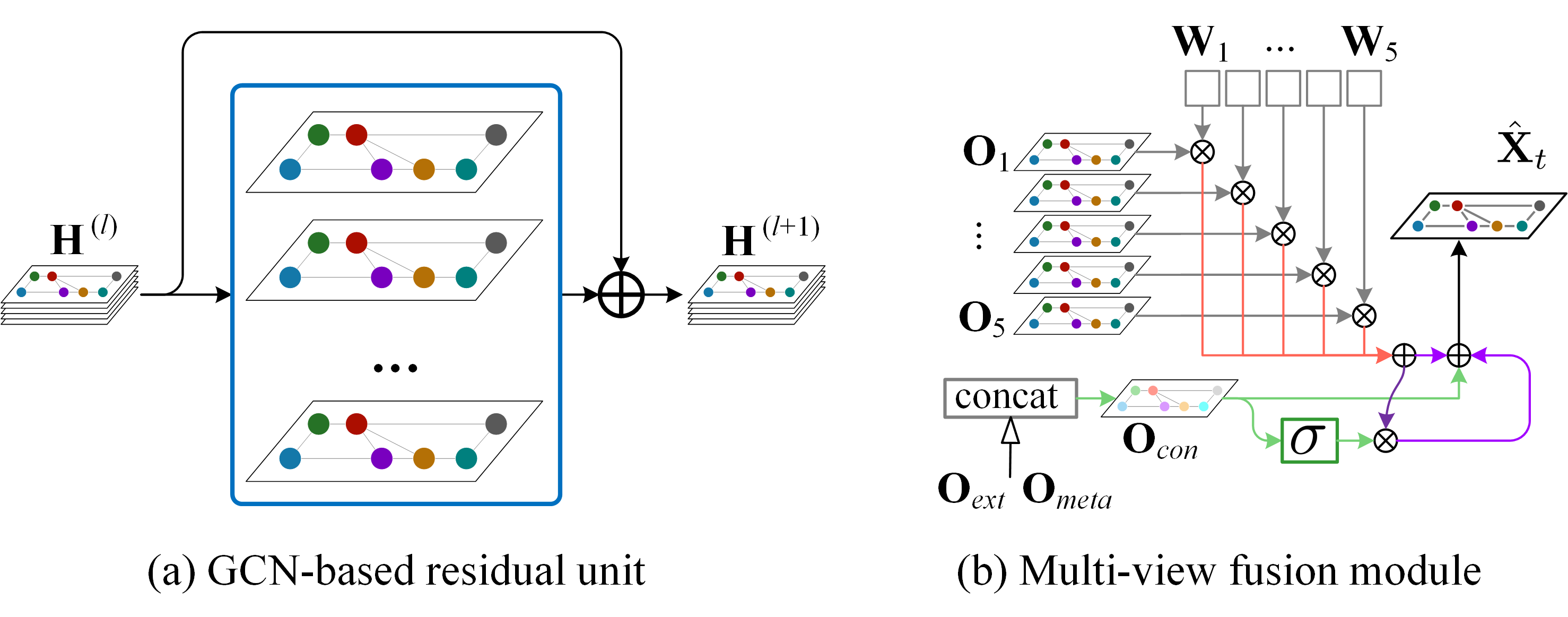}
	\caption{Main components of \model{}}
	\label{fig:gcn_mvf}
\end{figure}

\noindent\textbf{GCN-based Residual Unit}. 
To capture $M$-hop spatial correlations and interactions, we stack $M$ spatial graph convolutional layers, inspired by graph convolutions \cite{kipf2017semi}. When $M$ is large, we need a very deep network. Residual learning \cite{He2015apa} allows neural networks to have a super deep structure of 100 layers. Here we propose a GCN-based residual unit that integrates the graph convolutional layer into the residual framework (Fig~\ref{fig:gcn_mvf}(a)). Formally, the residual unit is defined as: 
\begin{eqnarray}
\mathbf H^{(l+1)} = \mathbf H^{(l)} + f\left( {\textbf{Q}}^{-\frac{1}{2}} \tilde{\textbf{S}} {\textbf{Q}}^{-\frac{1}{2}} \textbf{H}^{(l)} \W^{(l)}\right)
\end{eqnarray} 
where $f$ is an activation function. 

By stacking multiple GCN-based residual units, we can build very deep neural networks to capture multi-hop spatial dependencies. 

\subsection{Multi-view Fusion}\label{sec:mvf}
We propose a \textit{multi-view fusion} (see Fig~\ref{fig:gcn_mvf}(b)) method to fuse the latent representations of many flow views with two global views (external and meta data). 
In our previous crowd flow prediction task \cite{zhang2017aaai}, we show that different regions have different temporal properties, but the degrees of influence may be different. Inspired by this, we here also employ the parametric-matrix-based fusion method \cite{zhang2017aaai} to fuse the outputs of five GCNs for temporal views as below
\begin{eqnarray}
\mathbf O = \mathbf W_1 \odot \mathbf O_1 + \mathbf W_2 \odot \mathbf O_2 + \cdots + \mathbf W_5 \odot \mathbf O_5
\end{eqnarray}
where $\mathbf W_1, \cdots, \mathbf W_5$ are the learnable parameters that adjust the degrees affected by closeness, daily period, weekly period, monthly trend, and quarterly trend, respectively. 

For the external factor $\textbf{I}_{ext}$ (like weather and holiday) and meta data $\textbf{I}_{meta}$ (\eg{} time of the day), we separately feed them into different fully-connected (FC) layers to obtain different latent representations $\O_{ext}$ and $\O_{meta}$. Then we simply concatenate all the outputs of the \textit{embed} module and add a FC layer following by reshaping, thereby obtaining $\mathbf O_{con} \in \mathbb R^{N\times C}$.  

\begin{figure}[!htbp]
	\centering
	\includegraphics[width=1.\linewidth]{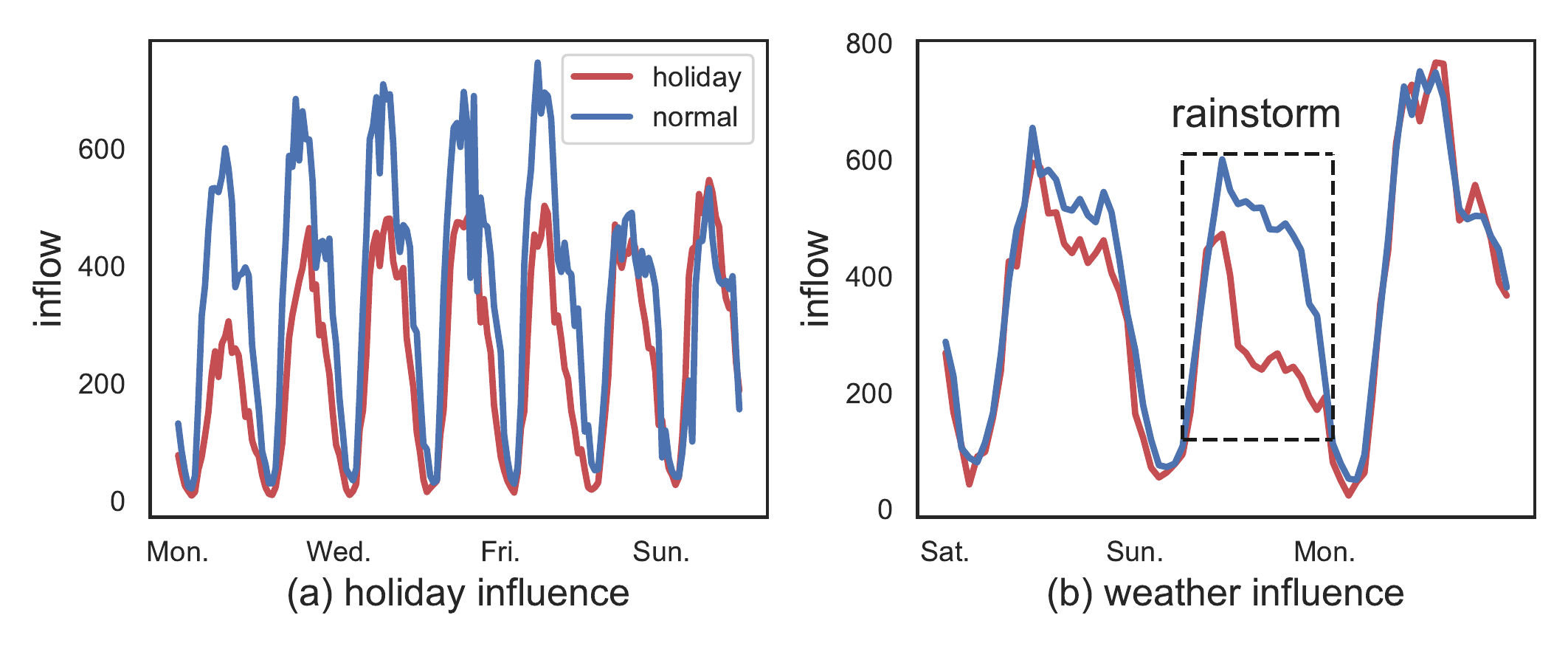}
	\caption{Different influences with external factors. (a) 2016 Chinese Spring Festival. (b) A stormy day vs. sunny day. The data is collected from TaxiBJ, as shown in Table~\ref{tab:ds}.}
	\label{fig:external}
\end{figure}

Different factors may change the flows in different ways. For example, holidays may moderate the crowd flows, as shown in Fig~\ref{fig:external}(a), while rainstorms may sharply and dramatically reduce the flows (Fig~\ref{fig:external}(b)). Specifically, the latter is just like a switch, changing flows tremendously change when it happens. On account of these insights, we leverage two different fusion methods to deal with these two types of situations. 
For the gradual changes, we propose employing a sum-fusion method, \eg{}, $\mathbf O_{con}$ + $\mathbf O$. For the sudden changes, we propose employing a gating-mechanism-based fusion, \eg{}, $\sigma(\mathbf O_{con})\odot \mathbf O$, where $\sigma$ is an approximated gating function such as $\sigmoid$. When the concatenated representation of $\mathbf O_{con}$ captures some special external information such as rainstorm weather, the term $\sigma(\mathbf O_{con})\odot \mathbf O$ will suddenly increase and become a much larger value due to the property of sigmoid function compared with $\mathbf O_{con}$. And in most common cases, this term should be close to zero without sudden changes.

Based on two fusion methods, the final output is calculated as
\begin{eqnarray}
\hat\X_t = f_{o}\left( \mathbf O + \mathbf O_{con} + \sigma(\mathbf O_{con})\odot \mathbf O\right) 
\end{eqnarray}
where $f_o$ is the activation function, \eg, $\tanh$, $\sigmoid$. 

\subsection{Loss and Algorithm}
Let $x$ and $\hat x$ be the observed and predicted values. 
The objective function we employ here is the Huber loss, which is an elegant compromise between squared-error loss $(x-\hat x)^2$ and absolute-error loss $|x-\hat x|$, and has been verified as a robust loss function for regression \cite{huber1964robust}. \\
The Huber loss, denoted $\mathcal L(x, \hat x)$, is defined by
\begin{equation}
\mathcal L(x, \hat x) = \left\{
\begin{aligned}
&\frac{1}{2} (x - \hat x)^2 & {\rm for}\ |x - \hat x| \leq \delta, \\
&\delta|x - \hat x| - \frac{1}{2}\delta^2,   & \rm{otherwise.}
\end{aligned}
\right.
\end{equation}
where $\delta$ is a threshold (1 by default). The Huber loss combines the desirable properties of squared-error loss near zero and absolute error loss when $|x - \hat x|$ is greater than $\delta$ (Fig~\ref{fig:gcn_effect} shows the empirical comparison). 


Let $\Theta$ be all the trainable parameters in \model. For the Huber loss it yields the following optimization problem,

\begin{equation}\label{eq:loss}
\argmin _\Theta \sum_{t\in \mathcal T}\sum_{i=1}^{N}\sum_{c=1}^{C} \mathcal L (\mathbf X_t[i,c], \hat {\mathbf X}_t[i,c])
\end{equation}
where $\mathbf X_t[i,c]$ means the element of the $i^{th}$ row and $c^{th}$ column of $\mathbf X_t$.

%% file: experiments.tex
\section{Experiments}
\subsection{Settings}
\noindent\textbf{Datasets}. We use four different datasets as shown in Table~\ref{tab:ds}. The details are described as follows:

\textbf{TaxiNYC\footnote{\url{http://www.nyc.gov/html/tlc/html/about/trip_record_data.shtml}}}: The trajectory data is taxi GPS data for New York City (NYC) from 1st Jan. 2011 to 30th Jun. 2016. We partition NYC into 100 irregular regions based on the map segmentation method (Section~\ref{sec:region}), and build the graph according to transition flow and geographical distance between regions, then we calculate crowd flows like \cite{hoang2016fccf}. 

\textbf{TaxiBJ}: Trajectory data is the taxicab GPS data in Beijing from four time intervals: 1st Jul. 2013-30th Oct. 2013, 1st Mar. 2014-30th Jun. 2014, 1st Mar. 2015-30th Jun. 2015, 1st Nov. 2015-10th Apr. 2016. The graph construction and crowd flow calculation method in Beijing is the same as that of NYC.  

\textbf{BikeDC\footnote{\url{https://www.capitalbikeshare.com/system-data}}}: The data is taken from the Washington D.C. Bike System. Trip data includes: trip duration, start and end station IDs, start and end times. There are 472 stations in total. For each station, we get two types of flows, where the inflow is the number of checked-in bikes, and the outflow is the number of checked-out bikes. Since many stations have no data or very few records, we remove these stations and apply a cluster operation \cite{KarypisK99} to the remaining stations using the average flow of historical observations, to get 120 irregular regions. 
We construct the graph with transition flow and geographical distance between these regions. 

\textbf{BikeNYC\footnote{\url{https://www.citibikenyc.com/system-data}}}: The data is taken from the NYC Bike system from 1st Jul. 2013 to 31th Dec. 2016. There are 416 stations in total. We also remove unavailable bike stations, and cluster the remaining stations into 120 regions. The graph construction and the bike flow calculation method in NYC is same as that of DC. 

For all aforementioned four datasets, we choose data from the last four weeks as the \textit{test set}, all data before that as the \textit{training set}. 
We build the commuting network (\ie{} graph) via the geographical distance between stations or regions, which can be viewed as nodes in the graph. The stations each have geospatial positions. For the regions, we approximate using the geospatial position of the central location of the region. 

\begin{table*}[t]\fontsize{9}{12}\selectfont
	\centering
	\caption{Datasets. Holidays include adjacent weekends. WS: wind speed. Temp.: temperature. }
	\label{tab:ds}
	\begin{tabular}{l|llll}
		\hline
		Dataset                        & TaxiNYC        & TaxiBJ    & BikeDC         & BikeNYC              \\
		\hline
		Data type                      & Taxi trip      & Taxi GPS     & Bike rent      & Bike rent         \\
		Location                            & NYC            & Beijing    & D.C.           & NYC            \\
		Start time                      & 1/1/2011       & 7/1/2013     & 1/1/2011       & 7/1/2013         \\
		End time                      & 6/30/2016      & 4/10/2016    & 12/31/2016     & 12/31/2016         \\
		Time interval                    & 1 hour         & 1 hour    & 1 hour         & 1 hour             \\
		\hline
		\# timesteps       & 48192          & 12336   & 52608          & 30720                          \\
		\# regions (stations) & 100            & 100    & 120 (472)           & 120 (416)                           \\
		\hline
		\# holidays   & 627            & 105 & 686            & 401                              \\
		Weather      & \textbackslash & 16 types & \textbackslash & \textbackslash  \\
		Temp. / $^\circ$C     & \textbackslash & {[}-24.6,41{]}       & \textbackslash & \textbackslash       \\
		WS / mph       & \textbackslash & {[}0,48.6{]} & \textbackslash & \textbackslash    \\  
		\hline    
	\end{tabular} 
\vspace*{-12pt}
\end{table*}


\noindent\textbf{Baselines.}
We compare \model with the following 9 models:
\begin{itemize}
		\vspace*{-0pt}\item \textbf{HA}: \textit{Historical average}, which models crowd flows as a seasonal process, and uses the average of previous seasons as the prediction with a period of one week. For example, the prediction for this Tuesday is the averaged crowd flows from all historical Tuesdays.  
		\vspace*{-0pt}\item \textbf{VAR}: \textit{Vector auto-regressive} is a more advanced spatio-temporal model, which is implemented using the \textit{statsmodel} python package\footnote{http://www.statsmodels.org}. The number of lags is set as 3, 5, 10, or 30. The best result is reported. 
	\vspace*{-0pt}\item \textbf{GBRT}: \textit{Gradient boosting decision tree} \cite{friedman2001greedy}. It uses the same features as the input of ANN. The optimal parameters are achieved by the grid search.
	\vspace*{-0pt}\item \textbf{FC-LSTM}: Encoder-decoder framework using LSTM \cite{srivastava2015unsupervised}. Both encoder and decoder have two recurrent layers with 128 or 64 LSTM units. 
	\vspace*{-0pt}\item \textbf{GCN}: We build a 3-layer supervised \textit{graph convolutional network} where the graph convolution \cite{kipf2017semi} is employed. The inputs are the previous 6 timesteps and the output is the target timestep. 
	\vspace*{-0pt}\item \textbf{DCRNN}: We build a 2-layer supervised \textit{diffusion convolutional recurrent neural network} \cite{lidiffusion}, which achieves state-of-the-art results on predicting traffic speed on roads. The inputs are the previous 6 timesteps and the output is the target timestep or timesteps. 
	\vspace*{-0pt}\item \textbf{FCCF}: Forecasting Citywide Crowd Flow model based on Gaussian Markov random fields \cite{hoang2016fccf}, that leverages flows in all individual regions and transitions between regions as well as external factors. As other baselines did not use the transition features, we remove the transition to get a new baseline, named \textbf{FCCFnoTrans}.  
	\vspace*{-0pt}\item \textbf{ST-MGCN}: Forecasting ride-hailing demand with spatiotemporal multi-graph convolution network.\cite{geng2019multigcn}. We reproduce the model referring the paper and using the recommended model settings in the paper.
\end{itemize}

\begin{table*}[!b]\fontsize{9}{12}\selectfont
	\centering
	\caption{Comparisons with baselines on four datasets based two metrics: RMSE and MAE (the smaller the better). HA and VAR are time-series models; GBRT use the spatial and temporal features; FC-LSTM/GCN/DCRNN/ST-MGCN are neural networks. FCCF/FCCFnoTran are based on Gaussian Markov random fields}
	\label{tab:results}
	{\begin{tabular}{c||c|cccccccccc}
		\hlinew{1pt}
	
	Dataset & Metric & HA     & VAR   & GBRT  & FC-LSTM & GCN   & DCRNN & FCCFnoTrans &  FCCF & ST-MGCN & \textbf{MVGCN} \\ \hline\hline
	\multirow{2}{*}{TaxiNYC}&RMSE  & 101.54 & 30.78 & 83.71 & 27.82   & 26.52 & 25.50 & 26.02 & 26.00    & 23.53 & \textbf{23.15} \\
	&MAE   & 33.02  & 11.21 & 23.46 & 11.25  & 11.12 & 11.20 & 9.25  & \textbf{9.24}  & 9.52   & 9.40 \\\hline\hline
	\multirow{2}{*}{\rotatebox{0}{TaxiBJ}} &RMSE  & 38.77  & 18.79 & 33.89 & 19.04   & 17.38 & 16.44 & 18.70 & 18.42  &  16.30  & \textbf{14.37} \\
	&MAE   & 22.89  & 11.38 & 20.34 & 11.86   & 10.60 & 9.68 & 10.74 & 10.44  & 10.18  & \textbf{9.11}  \\\hline\hline
	\multirow{2}{*}{BikeDC}&RMSE  & 2.61   & 1.95  & 3.46  & 1.88    & 1.88  & 1.90  & 2.22  & 2.14  & -   & \textbf{1.72}  \\
	&MAE   & 1.48   & 1.20  & 1.98  & 1.10    & 1.08  & 1.20  & 1.34  & 1.27  & -   & \textbf{1.00}  \\\hline\hline
	\multirow{2}{*}{BikeNYC}&RMSE  & 6.77   & 4.21  & 8.57  & 4.66    & 5.06  & 4.35  & 4.41  & 4.19 & -    & \textbf{4.15 } \\
	&MAE   & 4.00   & 2.71  & 5.17  & 2.78    & 2.85  & 2.90  & 2.79  & 2.65  & -   & \textbf{2.60} \\
			\hlinew{1pt}
	\end{tabular}}
\end{table*}

The neural network based models are implemented using TensorFlow and trained via backpropagation and Adam \cite{Kingma2014apa} optimization. 

\noindent\textbf{Preprocessing.} 
The Min-Max normalization method is used to scale the data into the range $[-1, 1]$ or $[0,1]$. In the evaluation, we re-scale the predicted value back to the normal values, and compare them with ground truth data. For external factors, we use one-hot encoding to transform metadata (\ie{}, the day of the week, the time of the day), holidays and weather conditions into binary vectors, and use Min-Max normalization to scale the Temperature and Wind speed into the range $[0, 1]$. 

\noindent\textbf{Environmental settings \& Hyperparameters.}
Our model as well as most baselines are implemented using TensorFlow and the model training process is performed on two Tesla V100 GPUs with 64GB RAM and 16GB GPU memory. The training time varies from 30 minutes to 3 hours on different datasets.
The detailed hyperparameter settings about our model are as follows:
(1) For lengths of the five dependent sequences, we set them as: $l_r$, $l_d$, $l_w$, $l_m$, $l_q$ $\in \{0,\cdots, 6\}$. 
(2) The number of graph convolutional layers is set as $\{3, \cdots, 7\}$, no regularization is used. 
(3) The hidden unit is set as 10 for each embed layer by default. 
(4) The training data is split into three parts: the last four weeks' data is used as the test set, adjacent previous four weeks' data is used as validation set and the rest of the data is used to train the models. The validation set is used to control the training process by early stopping and choose our final model parameters for each model based on the best validation score. 
(5) The batch size is set as 32. 
(6) The learning rate is set as $0.0003$. 
(7) The training epoch is set as 1000, early stopping patience is set as 50.
For all trained models, we only select the model which has the best score on the validation set, and evaluate it on the test set. 

\noindent\textbf{Evaluation Metrics.} For the evaluation of ST-prediction, we employ two metrics: 
Root Mean Square Error (RMSE), Mean Absolute Error (MAE), both of which are widely used in the regression tasks. 
Given predicted values $\{\hat x_i\}$ and ground-truth values $\{x_i\}$, the RMSE and MAE are respectively calculated as below
\begin{eqnarray}
RMSE = \sqrt{\frac{1}{N} \sum_i (x_i - \hat x_i)^2}, \quad
MAE = {\frac{1}{N} \sum_i |x_i - \hat x_i|} \nonumber
\end{eqnarray}
where $N$ is the total number of all predicted values. 

\subsection{Comprehensive Results}
Table~\ref{tab:results} presents a comprehensive comparison with all 9 baselines. 
In general, it indicates that our \model{} performs best on all datasets based on two metrics except MAE on TaxiNYC. Comparing our \model{} with the state-of-the-art model ST-MGCN, both exploit the graph convolution technique. But our model aims to model different data views using the same graph, but ST-MGCN attempts to build multiple semantic graphs to make more accurate predictions. Due to the lack of graph data for ST-MGCN for bike datasets, we only report the performances of ST-MGCN on taxi datasets. We can find that our model performs better than ST-MGCN in two datasets on RMSE and MAE metrics.

Among four datasets, we can observe that our \model achieves the greatest improvement on the dataset TaxiBJ. This is because the TaxiBJ dataset contains more external information, like weather, temperature, and wind speed. 
We find that FCCF performs very well because it also considers the period and trend as well as external information, even the transitions between regions. 
When transition features are removed, FCCF is degraded into the model FCCFnoTrans, resulting in a small increase in both RMSE and MAE, which shows the effectiveness of transition features. FC-LSTM and DCRNN perform worse than FCCF and \model because they are used to model sequences and do not consider period and trend in the crowd flow data.



\subsubsection{Results on sudden changes}
\begin{figure}[!htbp]
	\centering
	\subfigure[{TaxiNYC}]{\includegraphics[width=.49\linewidth]{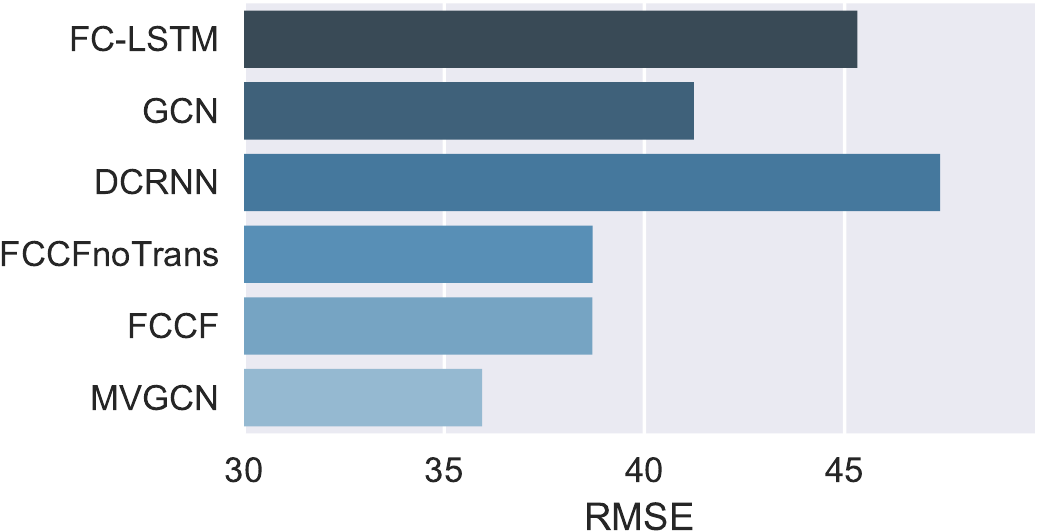}}
	\subfigure[{TaxiBJ}]{\includegraphics[width=.49\linewidth]{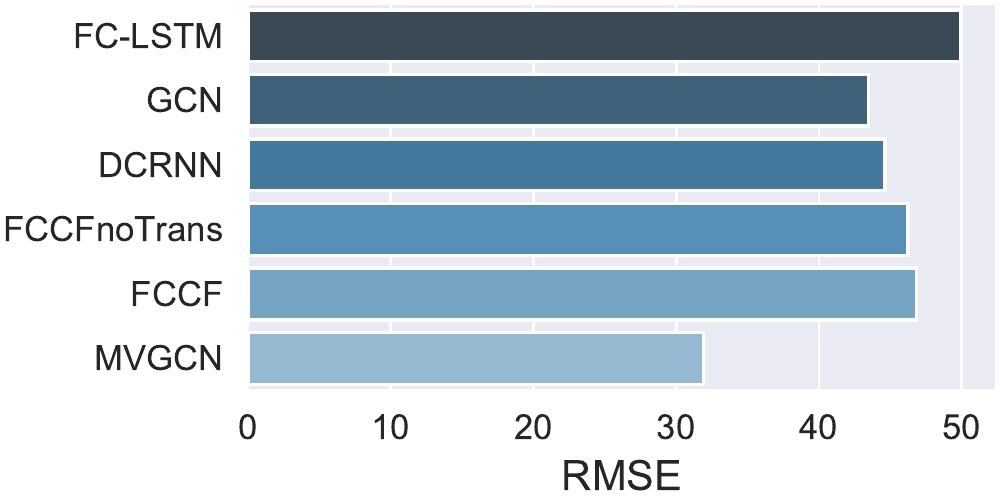}}
		\subfigure[{BikeDC}]{\includegraphics[width=.49\linewidth]{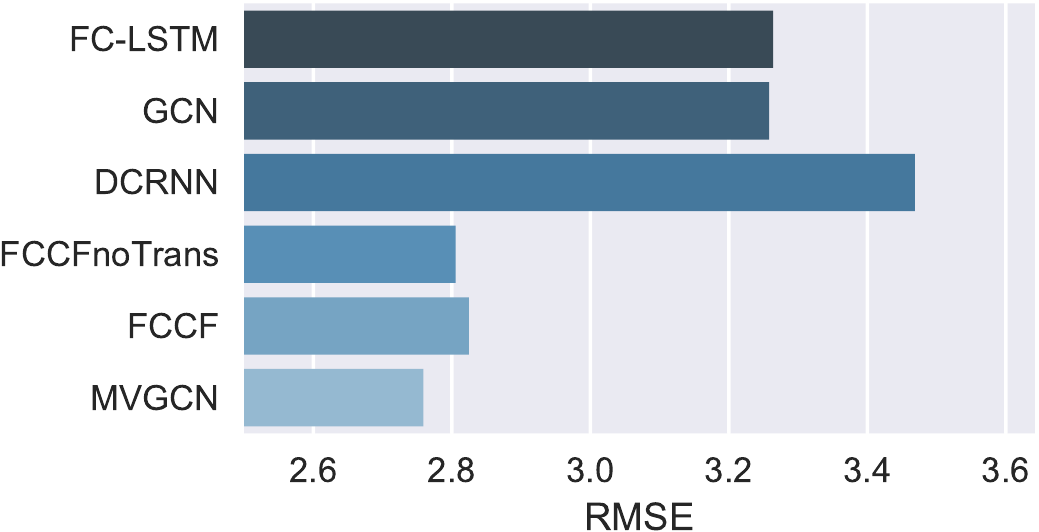}}
	\subfigure[{BikeNYC}]{\includegraphics[width=.49\linewidth]{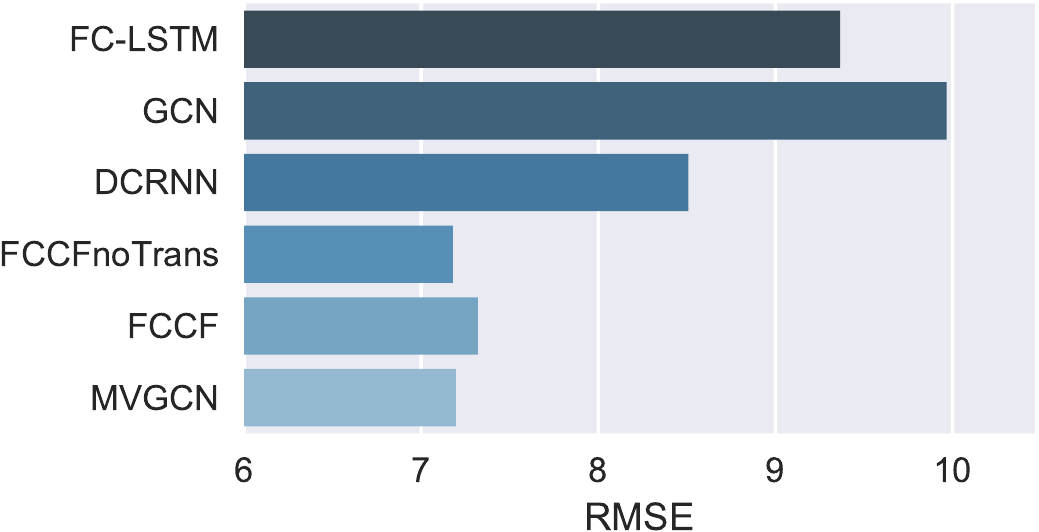}}
	\caption{RMSE comparisons on \textit{sudden changes} in the four datasets. }
	\label{fig:abnormal}
\end{figure}

\begin{figure}[!htbp]
 	\centering
	\includegraphics[width=.99\linewidth]{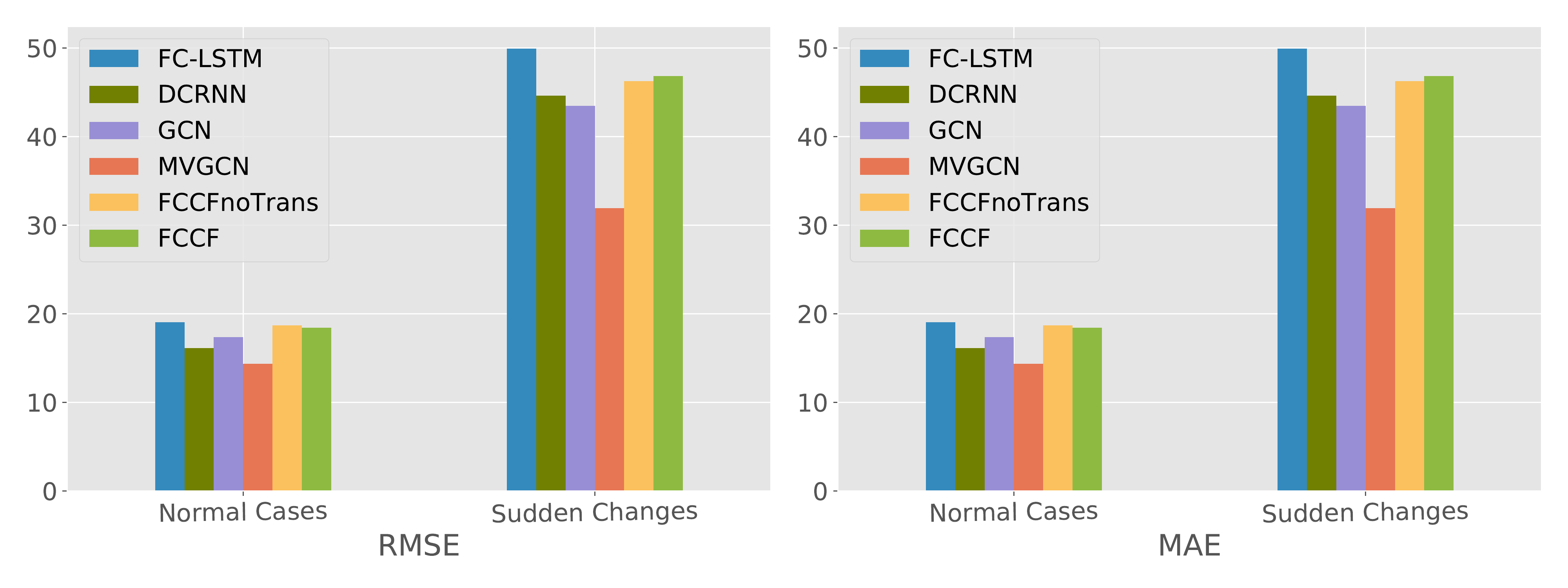}
	\caption{Performance comparison on normal cases and sudden changes in TaxiBJ dataset. }
	\label{fig:norm_abnormal}
\end{figure}

Fig~\ref{fig:abnormal} presents the comparisons between \model and the five baselines on sudden changes cases, which may be caused by anomalous weather or traffic events. 
For each timeslot $t$, we calculate the traffic flow difference with previous timeslot $t-1$ of all regions. Then we sort the absolute values of all differences in descending order and define the top $5\%$ as the timeslots where sudden changes happen. And the left $95\%$ timeslots are as normal cases.
We obverse that our \model greatly outperforms all other baselines, especially on TaxiBJ. As shown in Fig~\ref{fig:norm_abnormal}, our model performs better than baselines on both normal cases and sudden changes, besides, achieves more improvements on the latter.
One reason may be that our \model can effectively model weather data that is more complete in TaxiBJ. 

\subsubsection{Results on multi-step prediction}
\begin{figure}[!htbp]
	\centering
		\subfigure[{RMSE}]{\includegraphics[width=.49\linewidth]{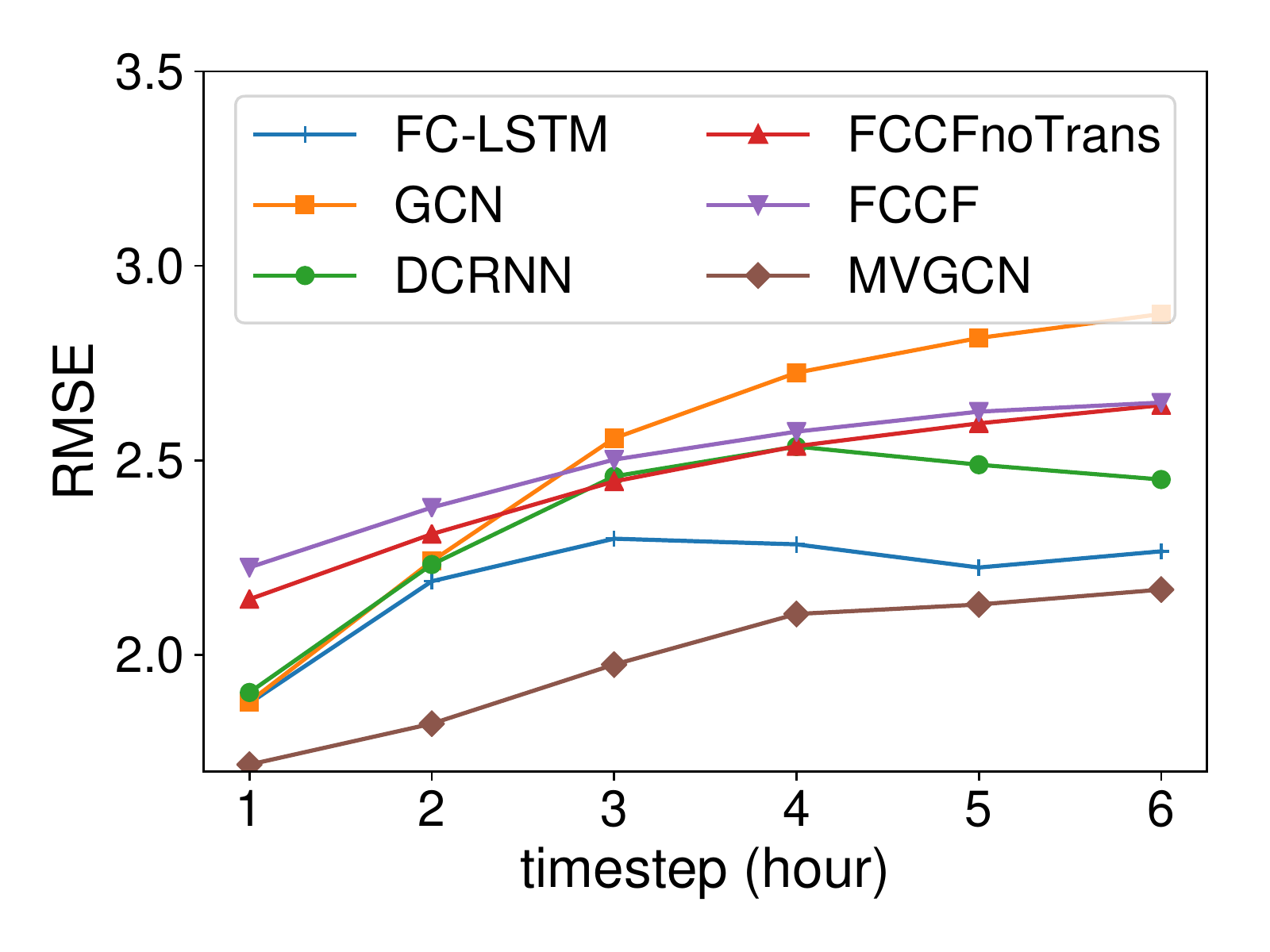}}
		\subfigure[{MAE}]{\includegraphics[width=.49\linewidth]{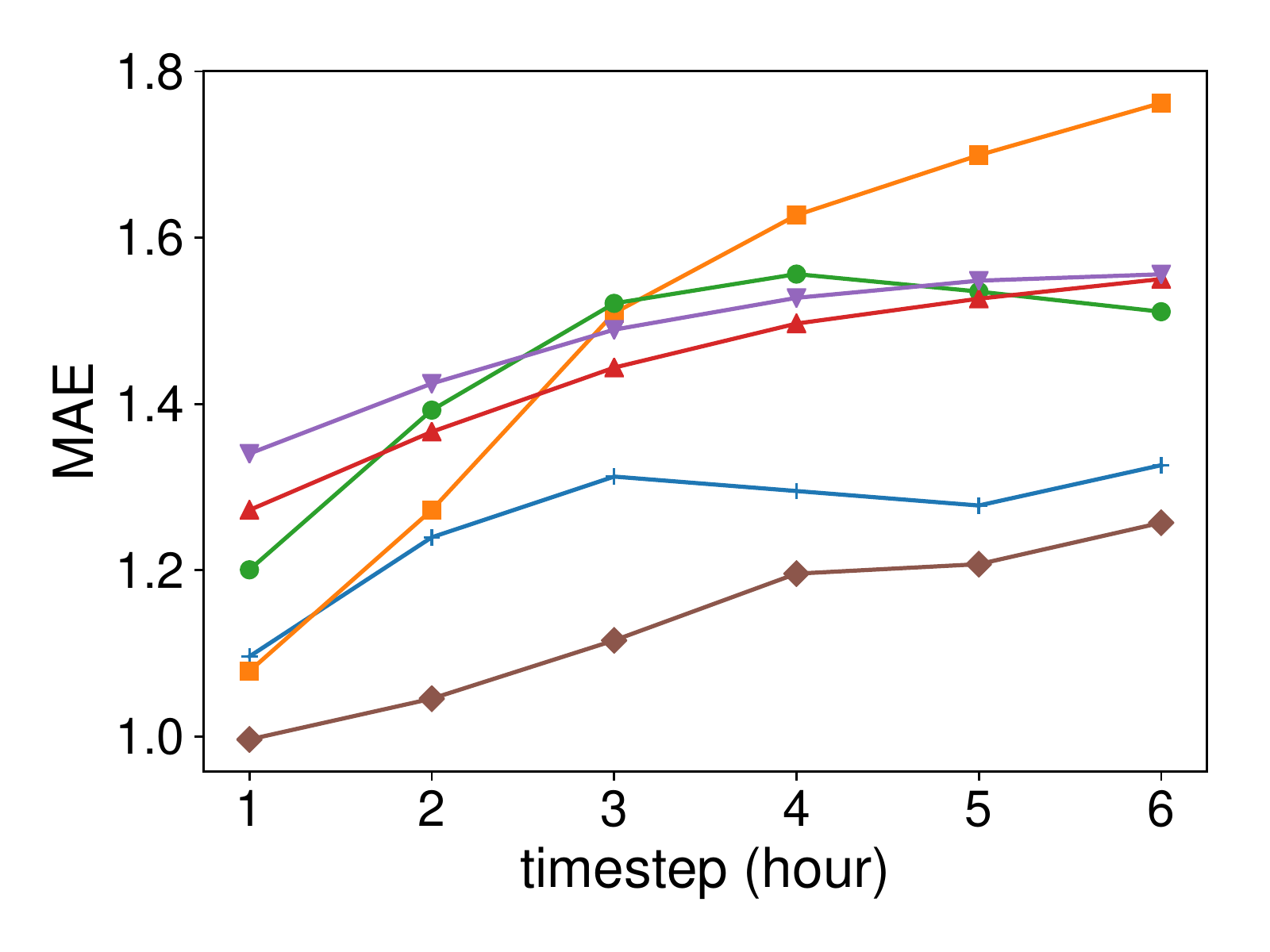}}
		\vspace*{-10pt}
	\caption{Step-wise comparisons on the BikeDC test set.}
	\label{fig:multi_steps}
\end{figure}

For further analysis, we present the multi-step prediction results based on RMSE and MAE over the dataset BikeDC in Fig~\ref{fig:multi_steps}. 
For the single-step prediction models, \eg{} our \model, we train different models for different timesteps. 
For the multi-step prediction models, including FC-LSTM and DCRNN, we use the previous 6 timesteps as the input sequence and the next 6 times as the target sequence, to train the model. Our \model is robust as the step number varies from 1 to 6, \ie{} small increase in both RMSE and MAE, achieving the best for all the 6 steps. We can observe that the original graph convolutional network (GCN) is not robust as the timestep increases, demonstrating that it does not work if we apply the existing models to the crowd flow prediction in a straightforward way. DCRNN performs less well because it also only use the sequence from the \textit{recent} timesteps, resulting in that it cannot capture period, trend, and external factors. 

\subsection{Effects of Different Components}

\subsubsection{Temporal view}
\begin{figure}[!htbp]
	\centering
	\subfigure[{RMSE}]{\includegraphics[width=.49\linewidth]{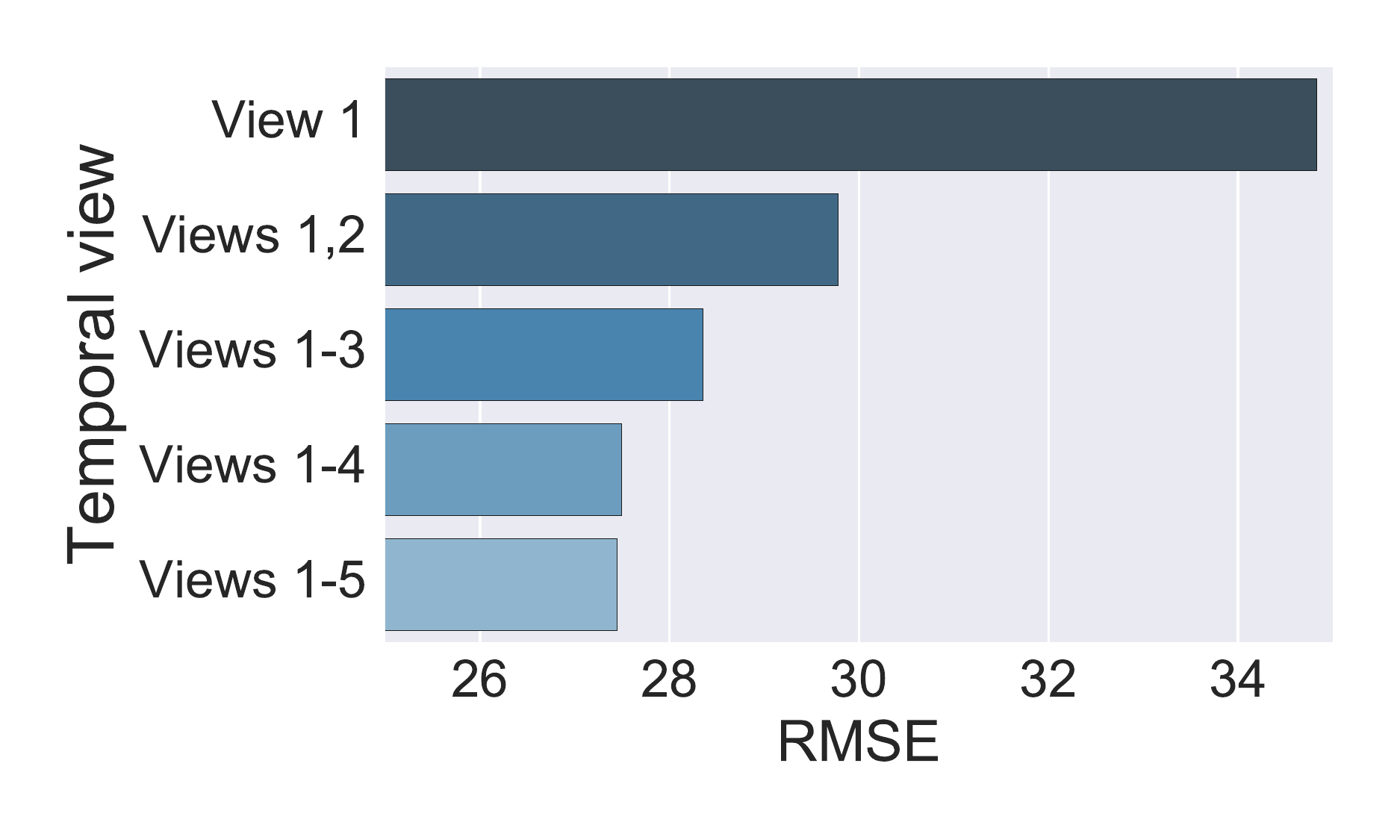}}
	\subfigure[{MAE}]{\includegraphics[width=.49\linewidth]{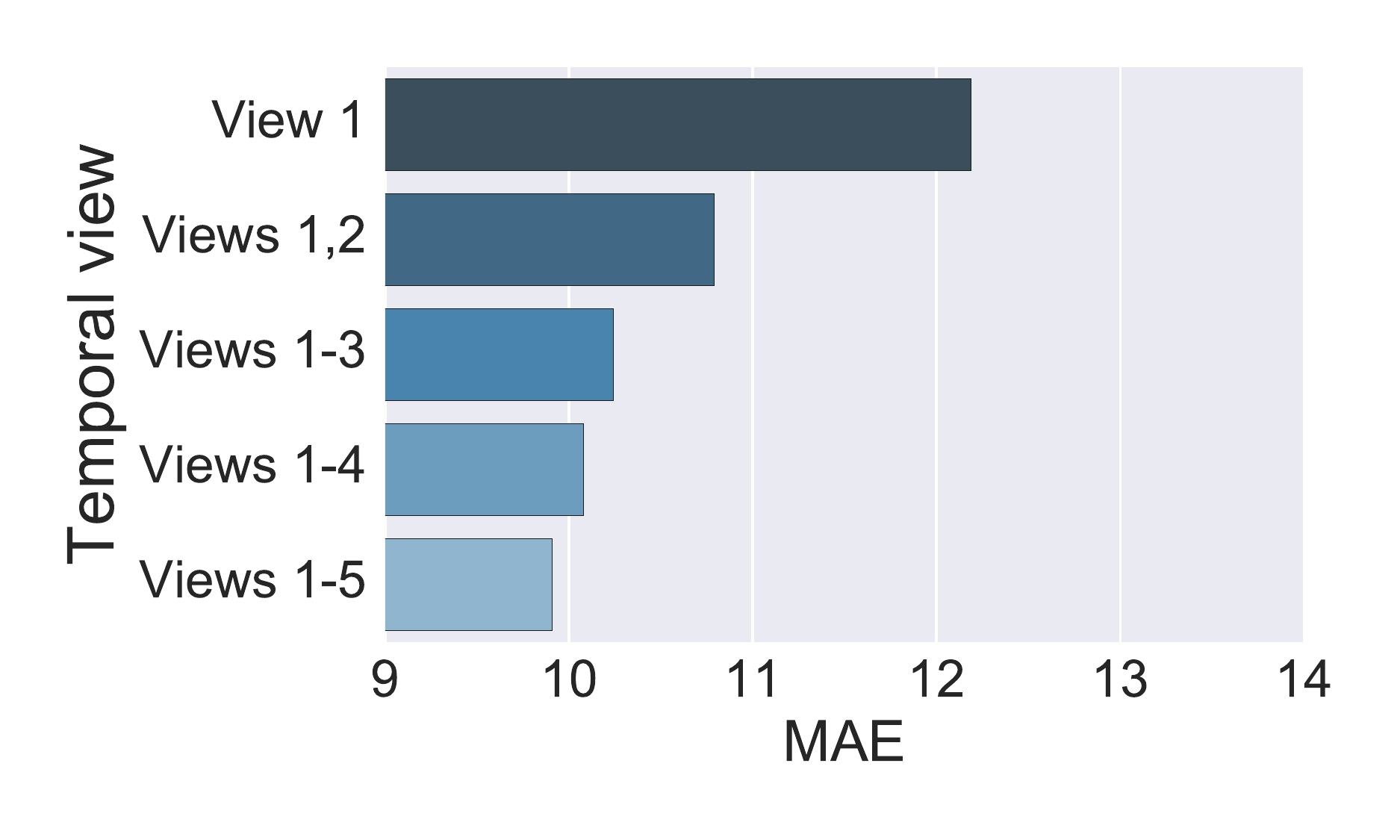}}
	\vspace*{-10pt}
	\caption{Effect of temporal views using TaxiNYC.}
	\label{fig:mvl}
\end{figure}

Fig~\ref{fig:mvl} demonstrates the different experiment effects of different combinations of temporal views based on RMSE and MAE, including recent (view 1), daily (view 2), weekly (view 3), monthly (view 4), and quarterly views (view 5). With only the recent view considered, we get a terrible result. When taking daily view into consideration, the result is greatly improved, indicating the periodicity is an important feature of traffic flow pattern. Also, the result becomes better and better with more temporal views considered. 

\subsubsection{Geospatial position.} 
Recall that in our model, we introduce a spatial graph convolution (see Eq.~\ref{eq:sc}), which integrates the geospatial position into the graph convolution. After eliminating such geospatial information, the layer is degraded as a graph convolution (Eq.~\ref{eq:gconv}). From Table~\ref{tab:others}, we observe that RMSE increases from 23.15 to 23.64 without the geospatial position, and MAE also becomes worse, demonstrating the effectiveness of the spatial graph convolution. 

\begin{table}[!htbp]\fontsize{9}{12}\selectfont
	\centering
	\vspace*{-10pt}
	\caption{Effect of different components on TaxiNYC test set. }
	\label{tab:others}
	\begin{tabular}{c|cc}
		\hlinew{1pt}
		Setting & RMSE         & MAE           \\
		\hline
		\model       & 23.15 & 9.40   \\
		\hline
		w/o geospatial position & 23.64 & 9.85  \\
		w/o external & 24.41 & 10.25  \\
		w/o metadata & 23.23 & 9.59  \\
		\hlinew{1pt}
	\end{tabular}
\end{table}

\subsubsection{Global information}
To show the effects of the \textit{embed} component, we compare the performance of \model{} under two settings: removing external factors or meta data, as shown in Table~\ref{tab:others}.
By eliminating the external factors, the RMSE increases from 23.15 to 24.41. Similarly, 
without the meta data, RMSE increases to 23.23. The results demonstrate that the external factors/meta data affect the prediction in an STG. 

\subsubsection{Huber loss and number of GCN layers.} 
\begin{figure}[!htbp]
	\centering
	\subfigure[{RMSE}]{\includegraphics[width=.49\linewidth]{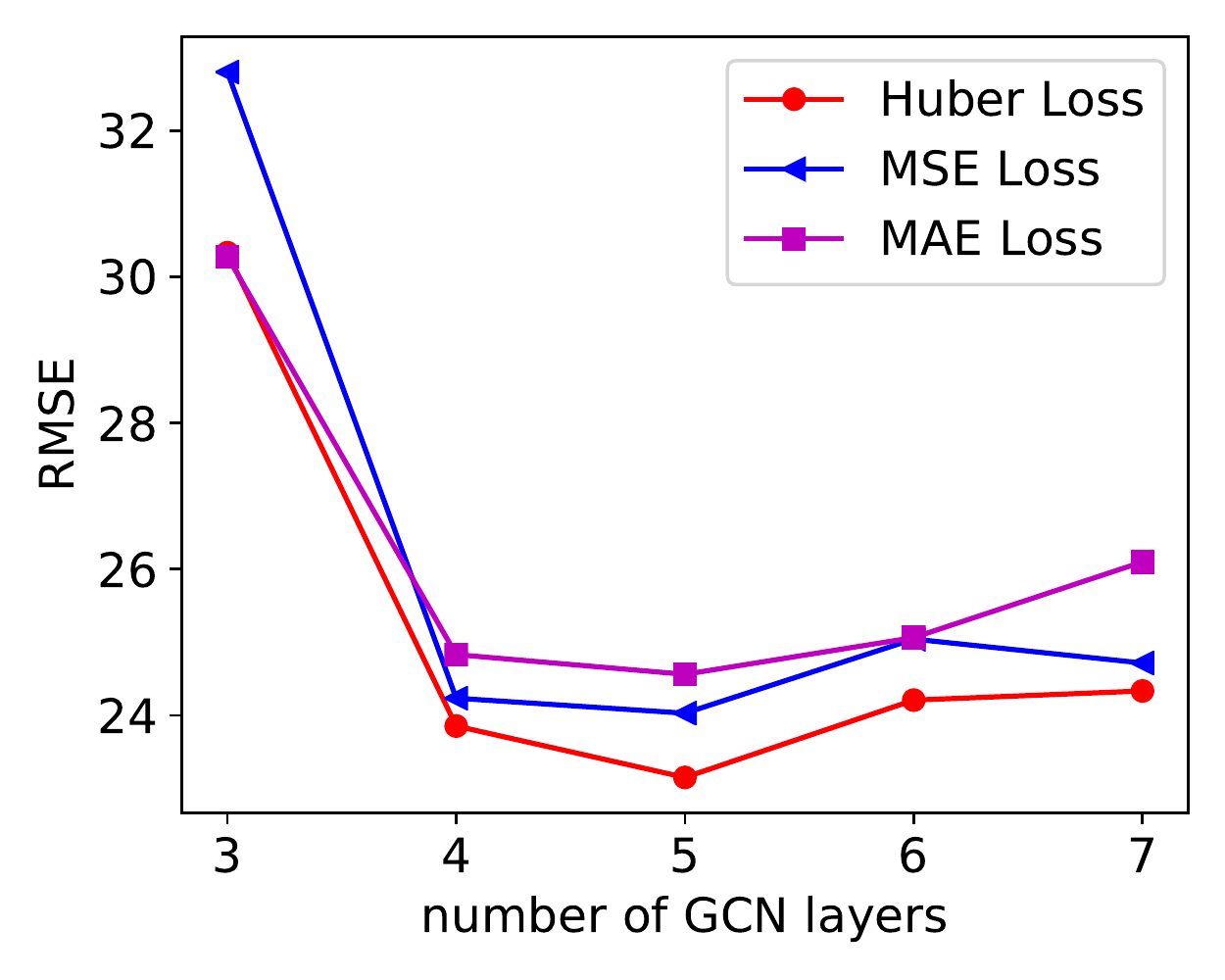}}
	\subfigure[{Training time}]{\includegraphics[width=.49\linewidth]{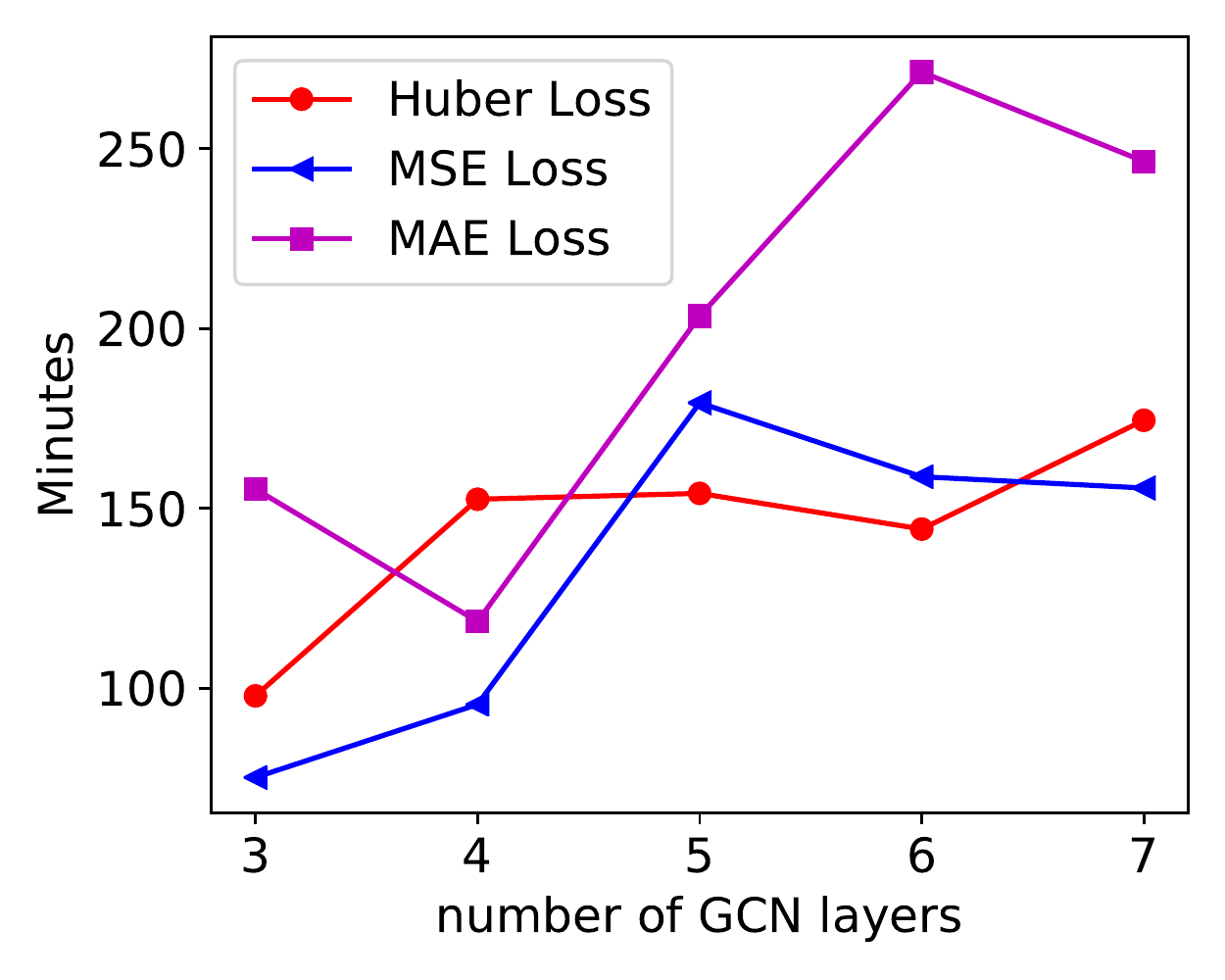}}
	\vspace*{-10pt}
	\caption{Model performance with varying number of GCN layers.}
	\label{fig:gcn_layers}
\end{figure}

\begin{figure}[!htbp]
	\centering
	\includegraphics[width=.99\linewidth]{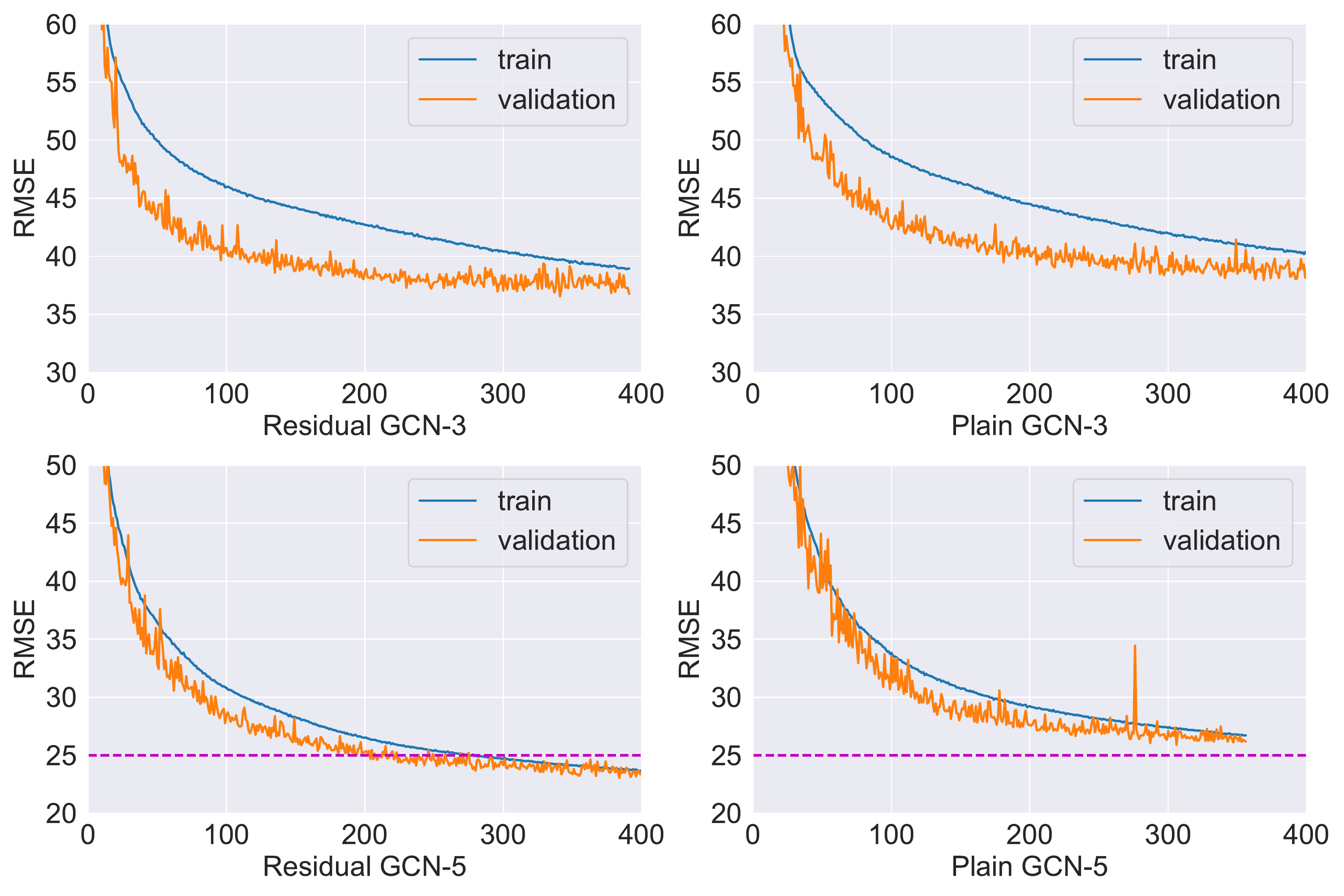}
	\caption{Performance curve on training and validation sets using plain or residual GCNs. The x-axis represents training epoch.}
	\label{fig:gcn_effect}
\end{figure}

To further investigate the effects of different loss functions and number of GCN layers. We perform some ablation studies and report results on TaxiNYC dataset with varying spatial graph convolutional layers or loss functions, as shown in Fig~\ref{fig:gcn_layers}.
We plot RMSE metric and can observe that the performance using RMSE, MAE or Huber as loss functions all first decrease and then increase as \textit{the number of GCN layers} increases. Best results occur when the number of GCN layers is 5. The figure demonstrates that deep networks yield better results but much deeper networks still cause the common problem of higher prediction error. And the training time when early stopping happens increases with the model depth on the whole.
To validate the effects of residual GCN layers, we compare the model with residual connections in GCN units with plain layers without residual connections. As shown in Fig~\ref{fig:gcn_effect}, we can observe that both of them perform similarly in shallow networks. But when the number of GCN layers is increased to 5, residual networks can achieve much better results, and both of them perform better than shallow networks while set at an appropriate depth.

%% file: relatedwork.tex

\section{Crowd flow forecasting system in irregular regions}
\begin{figure}[!htbp]
	\centering
	\subfigure[{Prediction VS Ground Truth for a certain region}]{\includegraphics[width=.99\linewidth]{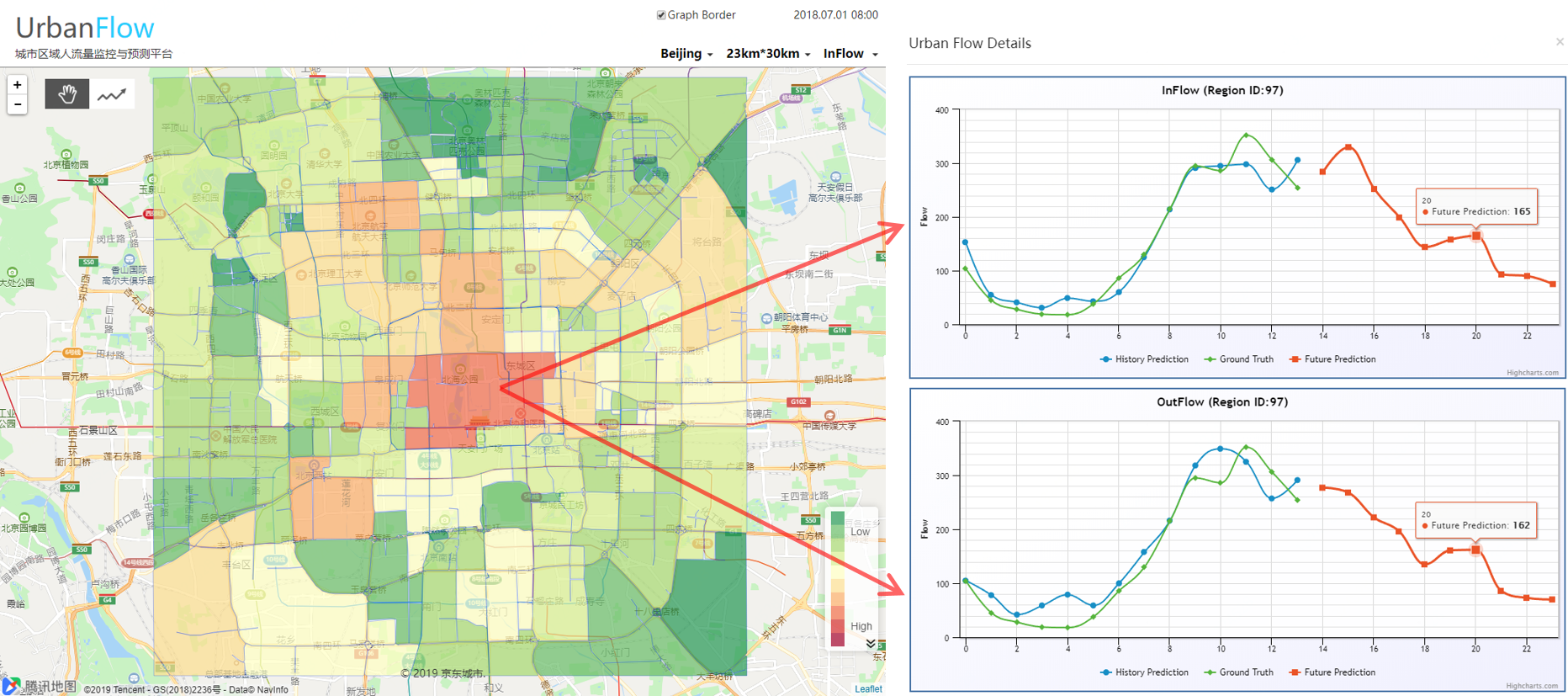} \label{fig:uf}}
	\subfigure[{Flow heatmap over time for the whole city}]{\includegraphics[width=.99\linewidth]{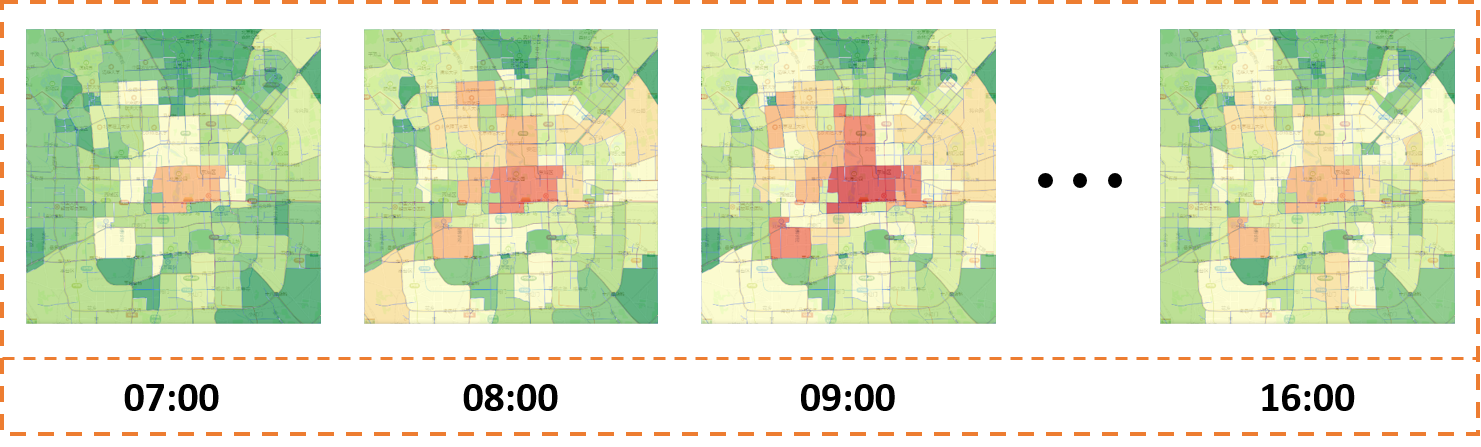} \label{fig:heatmap}}
	\caption{Web user interface overview of our UrbanFlow system. }
	\label{fig:urbanflow}
\end{figure}

We have developed a crowd flow forecasting demo (called UrbanFlow) in irregular regions internally, which can be accessed now\footnote{http://101.124.0.58/urbanflow\_graph}, as shown in Figure~\ref{fig:uf}. We have deployed it in the city area of Beijing, China, similar to that of our previous system for gridded regions. 
The detailed system architecture can be found in our previous work (Section 3 of \cite{zhang2017predicting}). 
Figure~\ref{fig:uf} shows the inflow and outflow results for a certain region in the system, where the green line represents the ground truth inflow or outflow in the previous 14 hours, the blue line denotes the prediction results in the 14 hours, and the orange line points the forecasting values in the next 10 hours. We can see the green and blue lines have very close values and similar trend, meaning that our \model can work effectively and well in the traffic flow forecasting system. 
Figure~\ref{fig:heatmap} displays another function view of overall flow changes of different time stamps for the whole city. We can observe the overall flow distribution varying with time. As the figure shows, in the morning rush hours, most regions have larger crowd flows because people are travelling from home, and the flows decrease in the mid-afternoon during which most people are working or resting indoors.

\section{Related Work}\label{sec:related_work}
\subsection{Spatio-Temporal Prediction}
There have been a lot of works about spatio-temporal prediction. Such as predicting travel speed and traffic volume on the road \cite{Wang2014TravelTE,Silva2015PotNAoS}. Most of them making predictions concerning single or multiple road segments, rather than citywide ones \cite{Xu2014IToITS,Chen2014}.
Recently, researchers have started to focus on city-scale traffic flow prediction \cite{hoang2016fccf,Yao2018DeepMS}. Specifically, \cite{hoang2016fccf} proposed a Gaussian Markov random field based model (called FCCF) that achieves state-of-the-art results on the crowd flow forecasting problem, which can be formulated as a prediction problem on an STG. \cite{Yao2018DeepMS} proposes a multi-view framework for citywide crowd flows prediction, but it is targeted for regular regions' flow prediction using of traditional convolutional neural networks.  And most spatiotemporal prediction works for raster-based data have been surveyed in \cite{wang2019stSurvey}. Work in \cite{geng2019multigcn} attempts to use multi-graph graph convolution to capture non-Euclidean correlation between regions, so they actually still perform their experiments in regular grid-based regions. Compared with this work, ours is targeted at the real problem of traffic prediction in irregular urban areas, and we also propose the method to process the traffic data and perform map segmentation with road networks.
\subsection{Classical Models for Time Series Prediction} 
Forecasting flow in a spatio-temporal network can be viewed as a time series prediction problem. Existing time-series models, like the auto-regressive integrated moving average model (ARIMA, \cite{box2015time}), seasonal ARIMA \cite{smith2002comparison}, and the vector autoregressive model \cite{chandra2009predictions} can capture temporal dependencies very well, yet it fails to capture spatial correlations. 
\vspace*{-8pt}
\subsection{Neural Networks for Sequence Prediction} 
Neural networks and deep learning \cite{LeCun2015DeepL} have achieved numerous successes in fields such as compute vision \cite{krizhevsky:2012imagenet,Simonyan2015VeryDC}, speech recognition \cite{graves2013speech}, and natural language understanding \cite{le2014distributed}.
Recurrent neural networks (RNNs) have been used successfully for sequence learning tasks \cite{sutskever2014sequence,Bahdanau2015NeuralMT}. The incorporation of long short-term memory (LSTM) \cite{hochreiter1997long} or gated recurrent unit (GRU) \cite{cho2014learning} enables RNNs to learn long-term temporal dependency. However, these neural network models can only capture spatial or temporal dependencies. Recently, researchers have combined the above networks and proposed a convolutional LSTM network \cite{xingjian2015convolutional} that learns spatial and temporal dependencies simultaneously but cannot be operated on spatio-temporal graphs. 
\cite{zhang2017aaai} proposed a spatio-temporal residual network, which is capable of capturing spatio-temporal dependencies as well as external factors in regular regions, yet it cannot be adapted to deal with graphs. 
\vspace*{-8pt}
\section{Conclusion}
We propose a novel multi-view deep learning model \model, consisting of several graph convolutional networks, to predict the inflow and outflow in each and every irregular region of a city. 
\model can not only capture spatial \textit{adjacent} and \textit{multi-hop} correlations as well as interactions, but also integrate the geospatial position via spatial graph convolutions. 
In addition, \model can capture many types of temporal properties, including closeness, periods (daily, weekly, \textit{etc}), and trends (\eg{} monthly, quarterly), as well as various external factors (like weather and event) and meta information (\eg{} time of the day). We evaluate our \model on four real-world datasets in different cities, achieving a performance which is significantly better than 8 baselines, including recurrent neural networks, and Gaussian Markov random field-based models. 

%% file: Paper.bbl
\begin{thebibliography}{10}
\providecommand{\url}[1]{#1}
\csname url@samestyle\endcsname
\providecommand{\newblock}{\relax}
\providecommand{\bibinfo}[2]{#2}
\providecommand{\BIBentrySTDinterwordspacing}{\spaceskip=0pt\relax}
\providecommand{\BIBentryALTinterwordstretchfactor}{4}
\providecommand{\BIBentryALTinterwordspacing}{\spaceskip=\fontdimen2\font plus
\BIBentryALTinterwordstretchfactor\fontdimen3\font minus
  \fontdimen4\font\relax}
\providecommand{\BIBforeignlanguage}[2]{{%
\expandafter\ifx\csname l@#1\endcsname\relax
\typeout{** WARNING: IEEEtranS.bst: No hyphenation pattern has been}%
\typeout{** loaded for the language `#1'. Using the pattern for}%
\typeout{** the default language instead.}%
\else
\language=\csname l@#1\endcsname
\fi
#2}}
\providecommand{\BIBdecl}{\relax}
\BIBdecl

\bibitem{Bahdanau2015NeuralMT}
D.~Bahdanau, K.~Cho, and Y.~Bengio, ``Neural machine translation by jointly
  learning to align and translate,'' \emph{CoRR}, vol. abs/1409.0473, 2015.

\bibitem{box2015time}
G.~E. Box, G.~M. Jenkins, G.~C. Reinsel, and G.~M. Ljung, \emph{Time series
  analysis: forecasting and control}.\hskip 1em plus 0.5em minus 0.4em\relax
  John Wiley \& Sons, 2015.

\bibitem{bruna2014spectral}
\BIBentryALTinterwordspacing
J.~Bruna, W.~Zaremba, A.~Szlam, and Y.~LeCun, ``Spectral networks and locally
  connected networks on graphs,'' in \emph{2nd International Conference on
  Learning Representations, {ICLR} 2014, Banff, AB, Canada, April 14-16, 2014,
  Conference Track Proceedings}, Y.~Bengio and Y.~LeCun, Eds., 2014. [Online].
  Available: \url{http://arxiv.org/abs/1312.6203}
\BIBentrySTDinterwordspacing

\bibitem{chandra2009predictions}
S.~R. Chandra and H.~Al-Deek, ``Predictions of freeway traffic speeds and
  volumes using vector autoregressive models,'' \emph{Journal of Intelligent
  Transportation Systems}, vol.~13, no.~2, pp. 53--72, 2009.

\bibitem{Chen2014}
P.-T. Chen, F.~Chen, and Z.~Qian, ``Road traffic congestion monitoring in
  social media with hinge-loss markov random fields,'' in \emph{2014 IEEE
  International Conference on Data Mining}.\hskip 1em plus 0.5em minus
  0.4em\relax IEEE, 2014, pp. 80--89.

\bibitem{cho2014learning}
K.~Cho, B.~van Merrienboer, Çaglar G{\"u}lçehre, D.~Bahdanau, F.~Bougares,
  H.~Schwenk, and Y.~Bengio, ``Learning phrase representations using rnn
  encoder-decoder for statistical machine translation,'' in \emph{EMNLP}, 2014,
  pp. 1724--1734.

\bibitem{defferrard2016convolutional}
M.~Defferrard, X.~Bresson, and P.~Vandergheynst, ``Convolutional neural
  networks on graphs with fast localized spectral filtering,'' in
  \emph{Advances in Neural Information Processing Systems}, 2016, pp.
  3844--3852.

\bibitem{friedman2001greedy}
J.~H. Friedman, ``Greedy function approximation: a gradient boosting machine,''
  \emph{Annals of statistics}, vol.~29, pp. 1189--1232, 2001.

\bibitem{geng2019multigcn}
X.~Geng, Y.~Li, L.~Wang, L.~Zhang, Q.~Yang, J.~Ye, and Y.~Liu, ``Spatiotemporal
  multi-graph convolution network for ride-hailing demand forecasting,'' in
  \emph{AAAI 2019.}, 2019, pp. 3656--3663.

\bibitem{graves2013speech}
A.~Graves, A.-r. Mohamed, and G.~Hinton, ``Speech recognition with deep
  recurrent neural networks,'' in \emph{2013 IEEE international conference on
  acoustics, speech and signal processing}.\hskip 1em plus 0.5em minus
  0.4em\relax IEEE, 2013, pp. 6645--6649.

\bibitem{hammond2011wavelets}
D.~K. Hammond, P.~Vandergheynst, and R.~Gribonval, ``Wavelets on graphs via
  spectral graph theory,'' \emph{Applied and Computational Harmonic Analysis},
  vol.~30, no.~2, pp. 129--150, 2011.

\bibitem{He2015apa}
K.~He, X.~Zhang, S.~Ren, and J.~Sun, ``Deep residual learning for image
  recognition,'' \emph{2016 IEEE Conference on Computer Vision and Pattern
  Recognition (CVPR)}, pp. 770--778, 2016.

\bibitem{hoang2016fccf}
M.~X. Hoang, Y.~Zheng, and A.~K. Singh, ``Fccf: forecasting citywide crowd
  flows based on big data,'' in \emph{Proceedings of the 24th ACM
  SIGSPATIAL}.\hskip 1em plus 0.5em minus 0.4em\relax ACM, 2016.

\bibitem{hochreiter1997long}
S.~Hochreiter and J.~Schmidhuber, ``Long short-term memory,'' \emph{Neural
  computation}, vol.~9, no.~8, pp. 1735--1780, 1997.

\bibitem{huber1964robust}
P.~J. Huber, ``Robust estimation of a location parameter,'' \emph{The Annals of
  Mathematical Statistics}, vol.~35, no.~1, pp. 73--101, 1964.

\bibitem{KarypisK99}
G.~Karypis and V.~Kumar, ``Parallel multilevel series k-way partitioning scheme
  for irregular graphs,'' \emph{{SIAM} Review}, vol.~41, no.~2, pp. 278--300,
  1999.

\bibitem{Kingma2014apa}
\BIBentryALTinterwordspacing
D.~P. Kingma and J.~Ba, ``Adam: {A} method for stochastic optimization,'' in
  \emph{3rd International Conference on Learning Representations, {ICLR} 2015,
  San Diego, CA, USA, May 7-9, 2015, Conference Track Proceedings}, Y.~Bengio
  and Y.~LeCun, Eds., 2015. [Online]. Available:
  \url{http://arxiv.org/abs/1412.6980}
\BIBentrySTDinterwordspacing

\bibitem{kipf2017semi}
T.~N. Kipf and M.~Welling, ``Semi-supervised classification with graph
  convolutional networks,'' in \emph{ICLR}, 2017.

\bibitem{krizhevsky:2012imagenet}
A.~Krizhevsky, I.~Sutskever, and G.~E. Hinton, ``Imagenet classification with
  deep convolutional neural networks,'' in \emph{Advances in neural information
  processing systems}, 2012, pp. 1097--1105.

\bibitem{le2014distributed}
Q.~V. Le and T.~Mikolov, ``Distributed representations of sentences and
  documents.'' in \emph{ICML}, vol.~14, 2014, pp. 1188--1196.

\bibitem{LeCun2015DeepL}
Y.~LeCun, Y.~Bengio, and G.~E. Hinton, ``Deep learning,'' \emph{Nature}, vol.
  521, pp. 436--444, 2015.

\bibitem{lidiffusion}
Y.~Li, R.~Yu, C.~Shahabi, and Y.~Liu, ``Diffusion convolutional recurrent
  neural network: Data-driven traffic forecasting,'' 2018.

\bibitem{lin2018air}
\BIBentryALTinterwordspacing
Y.~Lin, N.~Mago, Y.~Gao, Y.~Li, Y.-Y. Chiang, C.~Shahabi, and J.~L. Ambite,
  ``Exploiting spatiotemporal patterns for accurate air quality forecasting
  using deep learning,'' in \emph{SIGSPATIAL 2018}.\hskip 1em plus 0.5em minus
  0.4em\relax New York, NY, USA: ACM, 2018, pp. 359--368. [Online]. Available:
  \url{http://doi.acm.org/10.1145/3274895.3274907}
\BIBentrySTDinterwordspacing

\bibitem{shuman2013emerging}
D.~I. Shuman, S.~K. Narang, P.~Frossard, A.~Ortega, and P.~Vandergheynst, ``The
  emerging field of signal processing on graphs: Extending high-dimensional
  data analysis to networks and other irregular domains,'' \emph{IEEE Signal
  Processing Magazine}, vol.~30, no.~3, pp. 83--98, 2013.

\bibitem{Silva2015PotNAoS}
R.~Silva, S.~M. Kang, and E.~M. Airoldi, ``Predicting traffic volumes and
  estimating the effects of shocks in massive transportation systems,''
  \emph{Proceedings of the National Academy of Sciences}, vol. 112, no.~18, pp.
  5643--5648, 2015.

\bibitem{Simonyan2015VeryDC}
K.~Simonyan and A.~Zisserman, ``Very deep convolutional networks for
  large-scale image recognition,'' \emph{CoRR}, vol. abs/1409.1556, 2015.

\bibitem{smith2002comparison}
B.~L. Smith, B.~M. Williams, and R.~K. Oswald, ``Comparison of parametric and
  nonparametric models for traffic flow forecasting,'' \emph{Transportation
  Research Part C: Emerging Technologies}, vol.~10, no.~4, pp. 303--321, 2002.

\bibitem{srivastava2015unsupervised}
\BIBentryALTinterwordspacing
N.~Srivastava, E.~Mansimov, and R.~Salakhutdinov, ``Unsupervised learning of
  video representations using lstms,'' in \emph{Proceedings of the 32nd
  International Conference on Machine Learning, {ICML} 2015, Lille, France,
  6-11 July 2015}, ser. {JMLR} Workshop and Conference Proceedings, F.~R. Bach
  and D.~M. Blei, Eds., vol.~37.\hskip 1em plus 0.5em minus 0.4em\relax
  JMLR.org, 2015, pp. 843--852. [Online]. Available:
  \url{http://proceedings.mlr.press/v37/srivastava15.html}
\BIBentrySTDinterwordspacing

\bibitem{sutskever2014sequence}
I.~Sutskever, O.~Vinyals, and Q.~V. Le, ``Sequence to sequence learning with
  neural networks,'' in \emph{Advances in neural information processing
  systems}, 2014, pp. 3104--3112.

\bibitem{tobler1970computer}
W.~R. Tobler, ``A computer movie simulating urban growth in the detroit
  region,'' \emph{Economic geography}, vol.~46, no. sup1, pp. 234--240, 1970.

\bibitem{wang2019stSurvey}
\BIBentryALTinterwordspacing
S.~Wang, J.~Cao, and P.~S. Yu, ``Deep learning for spatio-temporal data mining:
  {A} survey,'' \emph{CoRR}, vol. abs/1906.04928, 2019. [Online]. Available:
  \url{http://arxiv.org/abs/1906.04928}
\BIBentrySTDinterwordspacing

\bibitem{wang2015deep}
W.~Wang, R.~Arora, K.~Livescu, and J.~Bilmes, ``On deep multi-view
  representation learning,'' in \emph{International Conference on Machine
  Learning}, 2015, pp. 1083--1092.

\bibitem{Wang2014TravelTE}
Y.~Wang, Y.~Zheng, and Y.~Xue, ``Travel time estimation of a path using sparse
  trajectories,'' in \emph{KDD}, 2014.

\bibitem{xingjian2015convolutional}
S.~Xingjian, Z.~Chen, H.~Wang, D.-Y. Yeung, W.-k. Wong, and W.-c. WOO,
  ``Convolutional lstm network: A machine learning approach for precipitation
  nowcasting,'' in \emph{Advances in Neural Information Processing Systems},
  2015, pp. 802--810.

\bibitem{xu2013survey}
\BIBentryALTinterwordspacing
C.~Xu, D.~Tao, and C.~Xu, ``A survey on multi-view learning,'' \emph{CoRR},
  vol. abs/1304.5634, 2013. [Online]. Available:
  \url{http://arxiv.org/abs/1304.5634}
\BIBentrySTDinterwordspacing

\bibitem{Xu2014IToITS}
Y.~Xu, Q.-J. Kong, R.~Klette, and Y.~Liu, ``Accurate and interpretable bayesian
  mars for traffic flow prediction,'' \emph{IEEE Transactions on Intelligent
  Transportation Systems}, vol.~15, no.~6, pp. 2457--2469, 2014.

\bibitem{yao2019revisiting}
H.~Yao, X.~Tang, H.~Wei, G.~Zheng, and Z.~Li, ``Revisiting spatial-temporal
  similarity: A deep learning framework for traffic prediction,'' in \emph{2019
  AAAI Conference on Artificial Intelligence (AAAI'19)}, 2019.

\bibitem{Yao2018DeepMS}
H.~Yao, F.~Wu, J.~Ke, X.~Tang, Y.~Jia, S.~Lu, P.~Gong, J.~Ye, and Z.~Li, ``Deep
  multi-view spatial-temporal network for taxi demand prediction,'' in
  \emph{AAAI}, 2018.

\bibitem{yuan2012}
J.~Yuan, Y.~Zheng, and X.~Xie, ``Discovering regions of different functions in
  a city using human mobility and pois,'' in \emph{The 18th {ACM} {SIGKDD}
  International Conference on Knowledge Discovery and Data Mining, {KDD}},
  2012, pp. 186--194.

\bibitem{yuan2012segmentation}
N.~J. Yuan, Y.~Zheng, and X.~Xie, ``Segmentation of urban areas using road
  networks,'' \emph{MSR-TR-2012--65, Technical Report}, 2012.

\bibitem{zhang2019multitask}
J.~{Zhang}, Y.~{Zheng}, J.~{Sun}, and D.~{Qi}, ``Flow prediction in
  spatio-temporal networks based on multitask deep learning,'' \emph{IEEE
  Transactions on Knowledge and Data Engineering}, pp. 1--1, 2019.

\bibitem{zhang2018gan}
\BIBentryALTinterwordspacing
J.~Zhang, X.~Shi, J.~Xie, H.~Ma, I.~King, and D.~Yeung, ``Gaan: Gated attention
  networks for learning on large and spatiotemporal graphs,'' in
  \emph{Proceedings of the Thirty-Fourth Conference on Uncertainty in
  Artificial Intelligence, {UAI} 2018, Monterey, California, USA, August 6-10,
  2018}, A.~Globerson and R.~Silva, Eds.\hskip 1em plus 0.5em minus 0.4em\relax
  {AUAI} Press, 2018, pp. 339--349. [Online]. Available:
  \url{http://auai.org/uai2018/proceedings/papers/139.pdf}
\BIBentrySTDinterwordspacing

\bibitem{zhang2017aaai}
J.~Zhang, Y.~Zheng, and D.~Qi, ``Deep spatio-temporal residual networks for
  citywide crowd flows prediction,'' in \emph{Thirty-First AAAI Conference on
  Artificial Intelligence}, 2017.

\bibitem{zhang2017predicting}
\BIBentryALTinterwordspacing
J.~Zhang, Y.~Zheng, D.~Qi, R.~Li, X.~Yi, and T.~Li, ``Predicting citywide crowd
  flows using deep spatio-temporal residual networks,'' \emph{Artif. Intell.},
  vol. 259, pp. 147--166, 2018. [Online]. Available:
  \url{https://doi.org/10.1016/j.artint.2018.03.002}
\BIBentrySTDinterwordspacing

\bibitem{zhao2017survey}
J.~Zhao, X.~Xie, X.~Xu, and S.~Sun, ``Multi-view learning overview: Recent
  progress and new challenges,'' \emph{Information Fusion}, vol.~38, pp. 43 --
  54, 2017.

\bibitem{Zheng2014AToISaTT}
Y.~Zheng, L.~Capra, O.~Wolfson, and H.~Yang, ``Urban computing: concepts,
  methodologies, and applications,'' \emph{ACM Transactions on Intelligent
  Systems and Technology (TIST)}, vol.~5, no.~3, p.~38, 2014.

\end{thebibliography}
